\documentclass[runningheads]{llncs}

\usepackage{eccv}

\usepackage{eccvabbrv}

\usepackage{graphicx}
\usepackage{booktabs}
\usepackage{multicol,multirow}

\usepackage[accsupp]{axessibility}  

\usepackage[breaklinks,colorlinks,citecolor=eccvblue]{hyperref}

\usepackage{orcidlink}

\begin{document}
\pdfobjcompresslevel=0

\title{VesselTok: Tokenizing Vessel-like 3D Biomedical Graph Representations for Reconstruction and Generation} 

\titlerunning{VesselTok}

\author{
    Chinmay~Prabhakar \inst{1,2} \orcidlink{0000-0002-1780-8108} \and
    Bastian Wittmann \inst{1,2} \orcidlink{0000-0002-6308-2926} \and
    Tamaz~Amiranashvili \inst{1,3} \orcidlink{0000-0001-8914-3427} \and
    Paul Büschl \inst{1} \orcidlink{0009-0002-6685-241X} \and 
    Ezequiel de la Rosa \inst{1}  \orcidlink{0000-0002-9042-1962} \and 
    Julian McGinnis \inst{3, 4}   \orcidlink{0009-0000-2224-7600}, \\
    Benedikt Wiestler \inst{3, 4} \orcidlink{0000-0002-2963-7772} \and
    Bjoern~Menze\thanks{Equal senior contribution} \inst{1,2} \orcidlink{0000-0003-4136-5690} \and
   Suprosanna~Shit$^\star$ \inst{1,2} \orcidlink{0000-0003-4435-7207} }

\authorrunning{Prabhakar et al.}
\institute{University of Zurich, Switzerland \and
ETH AI Center, Zurich, Switzerland \and
Technical University of Munich, Germany \and
Munich Center for Machine Learning (MCML), Germany\\
\email{chinmay.prabhakar@uzh.ch}}

\maketitle

\begin{abstract}
Spatial graphs provide a lightweight and elegant representation of curvilinear anatomical structures such as blood vessels, lung airways, and neuronal networks. Accurately modeling these graphs is crucial in clinical and (bio-)medical research.
However, the high spatial resolution of large networks drastically increases their complexity, resulting in significant computational challenges. In this work, we aim to tackle these challenges by proposing VesselTok, a framework that approaches spatially dense graphs from a parametric shape perspective to learn latent representations (tokens).
VesselTok leverages centerline points with a pseudo radius to effectively encode tubular geometry. Specifically, we learn a novel latent representation conditioned on centerline points to encode neural implicit representations of vessel-like, tubular structures. We demonstrate VesselTok's performance across diverse anatomies, including lung airways, lung vessels, and brain vessels, highlighting its ability to robustly encode complex topologies. To prove the effectiveness of VesselTok's learned latent representations, we show that they (i) generalize to unseen anatomies, (ii) support generative modeling of plausible anatomical graphs, and (iii) transfer effectively to downstream inverse problems, such as link prediction.

\end{abstract}

\section{Introduction}
\label{sec:intro}

\begin{figure}[t!]
    \centering
    \includegraphics[width=0.75\textwidth]{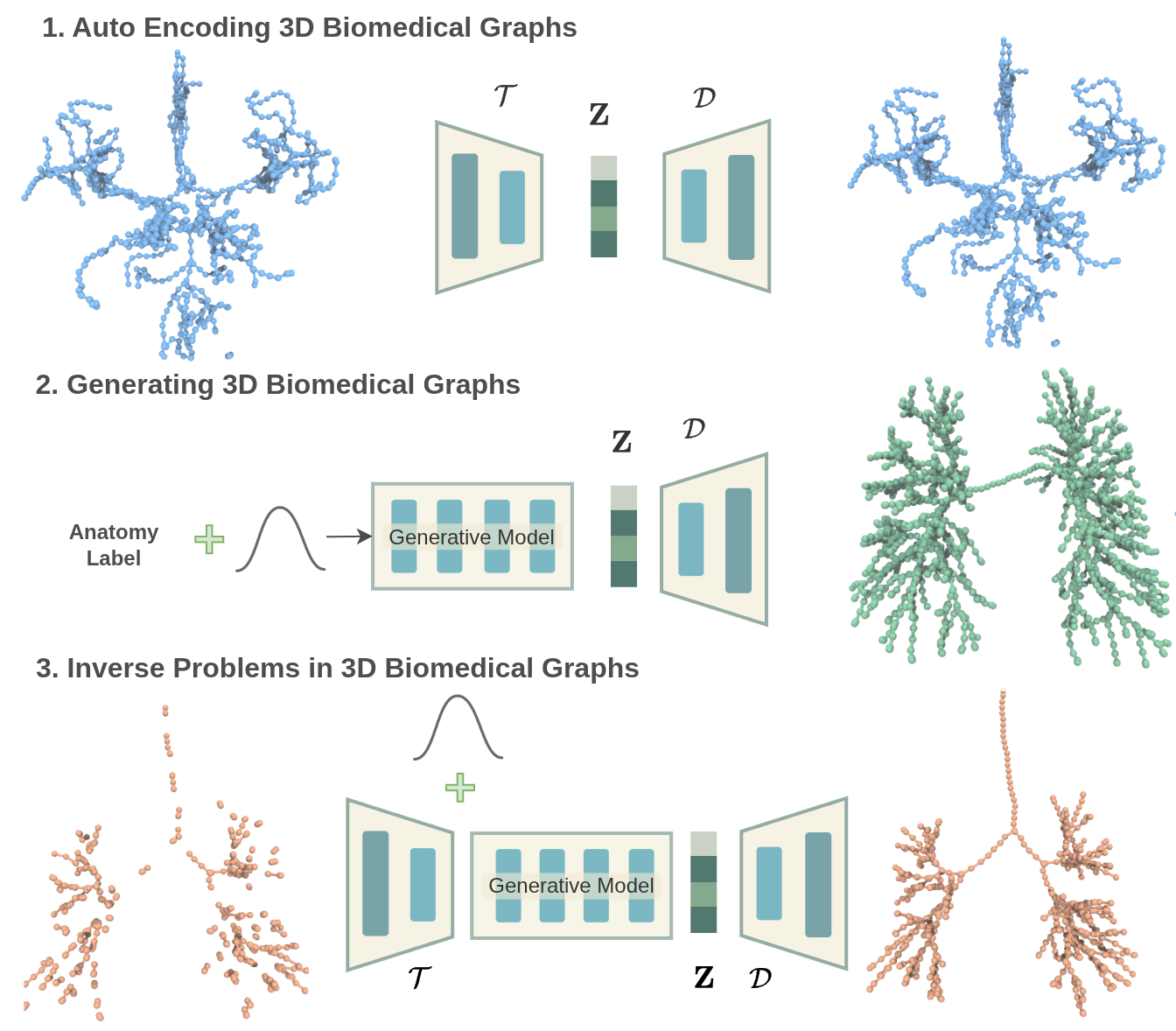}
    \caption{VesselTok provides expressive latent representations $\mathbf{Z}$ which can be effectively leveraged to several downstream tasks, including generalization to unseen anatomies, generative modeling (\ie, sample additional graphs), and inverse problems (\eg, repair of incomplete structures).}
    \label{fig:p1}
\end{figure}

Vessel-like, tubular networks are ubiquitous in physiological systems (\eg, blood vessel network, neuronal network, lung airways, or lymphatic network) and serve the purpose of connecting different anatomical regions, facilitating the transportation of signals and substrates. Studying their normal and abnormal characteristics is of immense importance in clinical and biomedical research, particularly in disease diagnosis, including cerebrovascular diseases~\cite{epp2023role, lin2022incomplete, wang2016skeleton}, pulmonary disorders~\cite{belchi2018lung, ortiz2023morphometric}, and diabetic peripheral neuropathy~\cite{dabbah2011automatic}. In this context, a 3D spatial graph representation comprised of the centerlines of these tubular structures efficiently encodes the necessary geometric and functional properties (lengths, radii, branch topology, and inputs for, \eg, computational flow modeling)~\cite{chapman2015automated,bumgarner2022open}. However, at high spatial resolution, these graphs typically contain a large number of nodes and edges, rendering them computationally challenging to process using off-the-shelf algorithms.
For this reason, previous graph modeling methods have only considered simplified structures~\cite{prabhakar20243d}, small components or subgraphs~\cite{feldman2023vesselvae}, and tree-like graphs~\cite{batten2025vector}. The objective of this work is to find \emph{high-fidelity} and \emph{computationally tractable} latent embeddings (tokens) of \emph{topologically complex, large 3D spatial graphs}, a shortcoming of existing techniques.

Although vessel-like structures exhibit substantial radius variation, we argue that explicitly modeling radius is not the main bottleneck for learning compact representations of complex tubular networks. In many anatomical systems (\eg, airways and cerebral vasculature), radii are anatomically constrained and vary smoothly along the centerline~\cite{kirst2020mapping,todorov2020machine}, making them amenable to reliably regress from centerline coordinates (see supplementary Sec.~B). We therefore assign a fixed pseudo-radius to centerline points and treat the true radius as an inferable attribute based on the structure. This choice is not meant to trivialize the task. Rather, it concentrates model capacity on encoding large-scale 3D geometry and topology while decoupling structural learning from the severe scale imbalance introduced by wide-radius distributions.

Recognizing the increased computational complexity from a pure graph perspective, we take an alternative approach and model the 3D spatial graph as a tubular surface (with a fixed pseudo-radius) represented as a continuous 3D occupancy field.
However, vascular networks differ from generic 3D shapes. They are sparse, thin, and highly branched, resulting in a high surface-to-volume ratio. Consequently, surface representations can contain far more samples than the underlying centerline. For example, a representative ATM case contains roughly 64,000 surface points but only 3,000 centerline points. Generic surface-point tokenizers~\cite{zhang20233dshape2vecset,zhao2025hunyuan3d} may therefore spend much of their fixed query budget on redundant tubular surface samples, while undersampling small branches, endpoints, and bifurcations. Therefore, we propose to operate on centerline points, allowing the model to focus its representational capacity on the intrinsic structure of high surface-to-volume ratio tubular networks rather than redundant surface geometry, resulting in more expressive latent representations.

Building on this intuition, we introduce VesselTok, a tokenizer that generates high-fidelity, yet computationally tractable, latent representations of vascular graphs. Our key insight is to represent a 3D graph as an implicit occupancy field and to tokenize its intrinsic centerline structure through an implicit decoder. This implicit representation enables a compact, expressive latent space that faithfully encodes the geometry and topology of large, complex vascular networks, making it suitable as a shared token space for reconstruction, generation, and infilling. VesselTok is trained on diverse datasets, resulting in a generalizable model that robustly encodes complex topologies. We further demonstrate the applicability of our generated 3D graph tokens in generative modeling and inverse problems, such as link prediction (see Fig.~\ref{fig:p1}).

\noindent
In summary, our contributions are as follows:
\begin{enumerate}
    \item We propose VesselTok, a state-of-the-art tokenizer for 3D spatial graphs that leverages the implicit representation of tubular shapes and is capable of generalizing to diverse curvilinear anatomical structures.
    \item We demonstrate downstream use of VesselTok's embeddings for unconditional and conditional generative modeling, 
    highlighting its representational strength to capture complex 3D topology.
    \item Further, we show VesselTok's generative prior in the token space can be used to solve inverse problems, such as link prediction.
\end{enumerate}
\section{Related Literature}
\label{sec:rel_lit}

\paragraph{Graph Tokenizer:}
Tokenizing graphs to learn semantically rich latent representations has been extensively explored. Pioneering works, such as VQGraph~\cite{yang2023vqgraph}, learn a VQ–VAE–style tokenizer that assigns each node’s local substructure to a discrete code, forming a codebook that captures diverse local patterns. Similarly, Liu et al.~\cite{liu2023rethinking} systematically study different tokenizers (node, edge, motif, and GNN-based) on molecules and show that subgraph-level tokenization, combined with an expressive decoder, substantially improves self-supervised representation learning. More recently, Wang et al.~\cite{wang2024learning} propose a Graph Quantized Tokenizer trained via multi-task self-supervised objectives and decoupled from the downstream Transformer. OpenGraph~\cite{xia2024opengraph} similarly introduces a unified tokenizer within a graph foundation model: graphs from diverse domains are mapped into a shared token space, which allows a single model to generalize to unseen graph properties and datasets.
However, all above described methods operate at native graph resolution, \ie, the latent retains the original node count. Hence, such methods are effective for downstream analysis but ill-suited to the computational demands of generating large graphs. In contrast, we explicitly prioritize compression, learning a compact latent that enables high-fidelity synthesis at scales far beyond prior approaches~\cite{vignac2023midi,prabhakar20243d}.

\paragraph{Shape Tokenizer:}
Recent efforts increasingly compress 3D shapes into compact tokens that serve as an interface to generative models. Early text–to–3D and image–to–3D pipelines, such as Shap-E~\cite{jun2023shap} and Meta 3D Gen~\cite {bensadoun2024meta}, demonstrated the effectiveness of latents for shape synthesis. Among other notable works, 3DShape2VecSet~\cite{zhang20233dshape2vecset} encodes surface point clouds into a set of latent vectors for diffusion-based shape modeling. Hunyuan3D 2.0~\cite{zhao2025hunyuan3d} proposes an improved architecture for shape encoding with important query point sampling. 3D Shape Tokenization~\cite{chang20243d} leverages continuous shape tokens learned via flow matching for image-to-3D and neural rendering. Dora~\cite{chen2025dora} addresses the sampling bias in Shape-VAEs through sharp-edge-aware sampling. Michelangelo~\cite{zhao2023michelangelo} aligns shape–image–text latents for conditional 3D generation. Other relevant work includes VAT~\cite{zhang20243d}, a variational tokenizer that compresses unordered 3D features into hierarchical latent tokens for autoregressive generation, Structured 3D Latents~\cite{xiang2025structured}, which uses an augmented sparse 3D grid to enable fast, high-quality text/image-conditioned 3D generation, and G3PT~\cite{zhang2024g3pt}, which uses transformer tokenizers for point clouds with occupancy decoders.
Although these methods aim for compact, faithful, cross-modal tokens, they are designed for generic shapes using surface-centric cues and omit vessel-specific, graph-oriented design. In contrast, VesselTok exploits the high surface-to-volume ratio of tubular networks by operating on centerline points.

\paragraph{Vessel Tokenizer:}
While early works explored generating vascular trees directly in uncompressed centerline space~\cite{kuipers2024generating,wolterink2018blood}, more recent approaches for tubular and tree-like anatomies have increasingly shifted toward learned latent representations and tokenized encodings, enabling more compact and scalable modeling. VesselVAE~\cite{feldman2023vesselvae} formulates a recursive variational autoencoder that maps a vessel tree to a compact latent vector. VesselGPT~\cite{feldman2025vesselgpt} builds a discrete codebook for vessels with a VQ-VAE applied to represent vascular structure as a short token sequence. Kuipers et al.~\cite{kuipers2025self} adapt the 3DShape2VecSet framework~\cite{zhang20233dshape2vecset} to vascular data, but evaluate it only on the relatively small Circle of Willis anatomy. Batten et al.~\cite{batten2025vector} propose to learn vector representations of vessel trees at two levels: a segment autoencoder that encodes branch centerlines, and a tree autoencoder that aggregates segment codes into a single vector. Similarly, Chen et al.~\cite{chen2025hierarchical} propose a hierarchical part-based model for generation. It first samples a global binary tree topology, then generates segment-level geometry, and finally assembles the full vessel tree by composing the synthesized segments according to the sampled topology. Beyond blood vessels, Zhang et al.~\cite{zhang2024implicit} utilize geometric correspondence in the segmentation space to encode airway trees using neural implicit representation. 
While these works take early steps toward vessel-specific latent modeling, they are largely limited to small geometries, tree-only structures, and simple vessel segments. In contrast, VesselTok scales to large geometries and supports complex topologies, including non-tree connectivity.

\section{Methodology}
\label{sec:meth}

\begin{figure*}[!t]
    \centering
    \includegraphics[width=0.95\linewidth, trim=0 0 200 0, clip]{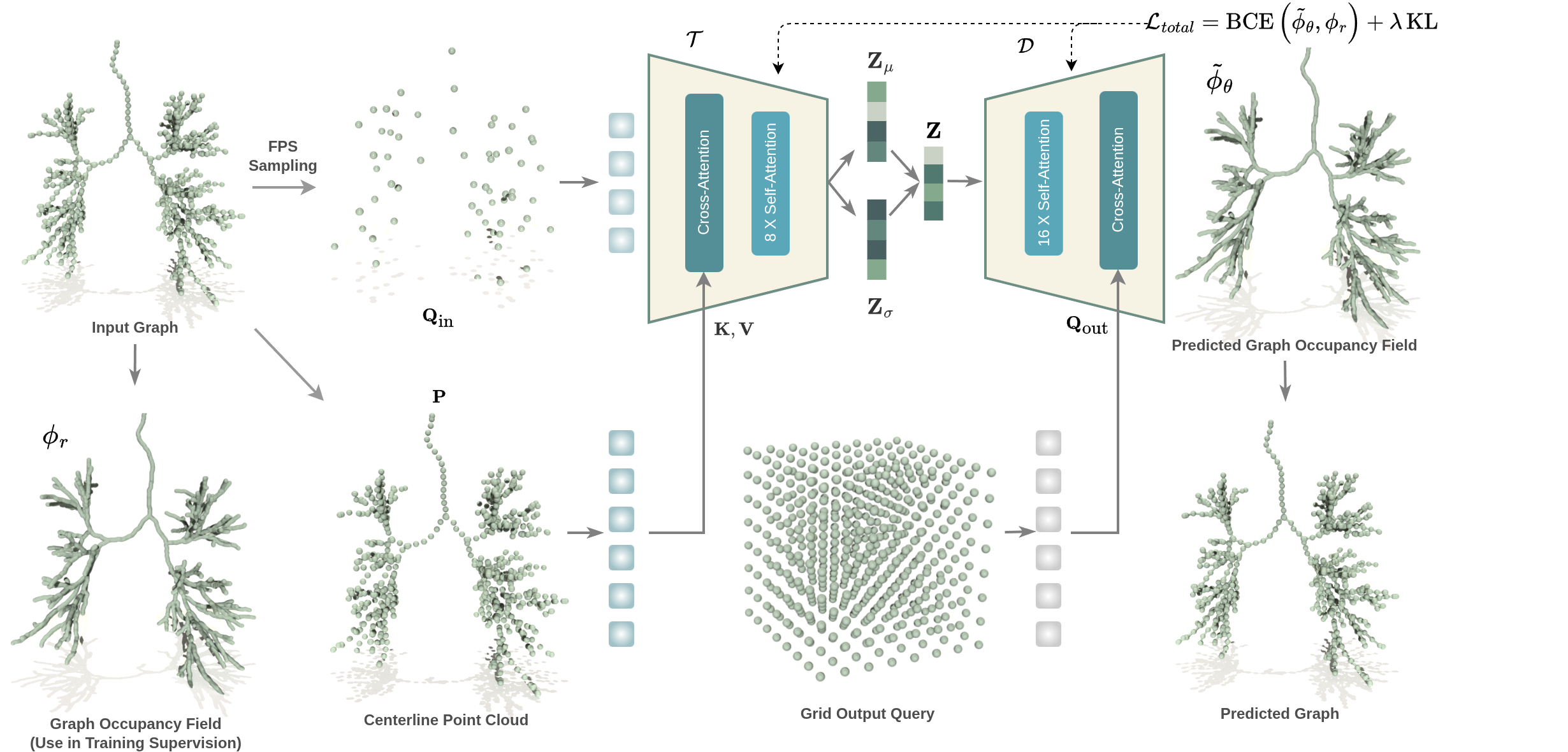}
    \caption{Architectural overview of VesselTok. VesselTok consists of an encoder $\mathcal{T}$, which extracts features from a pre-processed centerline point cloud $\mathbf{P}$ to generate a continuous, expressive, and compressed latent token $\mathbf{Z}$. A decoder $\mathcal{D}$ subsequently reconstructs the graph occupancy field $\Tilde{\phi}_\theta$ from $\mathbf{Z}$ via cross-attention to output query points $\mathbf{Q}_{\text{out}}$. The final graph is subsequently reconstructed from $\Tilde{\phi}_\theta$.}
    \label{fig:arch}
\end{figure*}

\subsection{Definitions}
\paragraph{Problem Statement:} Let \(G=(V, E, \mathbf{P})\) be a 3D spatial graph with vertices \(V=\{1,\dots,n\}\), edges \(E\subseteq V\times V\), and associated coordinates \(\mathbf{P}\in\mathbb{R}^{n\times 3}\). We aim to learn a graph tokenizer \(\mathcal{T}\) that maps \(G\) to a \(l\)-length token sequence with channel size $c$ (continuous tokens \(\mathbf{Z}\in\mathbb{R}^{l\times c}\) in case of VAE latents) together with a decoder \(\mathcal{D}\) such that the decoded graph \(\Tilde{G}:\mathcal{D}(\mathcal{T}(G))\) has the same topology and structure of the input graph $G$.

\paragraph{3D Graph Occupancy Field:} 
\label{sec:occupancy}

Since the position of the centerline points and the total number of points can vary without altering the underlying vessel structure (non-injective mapping), we aim to construct a robust latent representation of the graph structure. To this end, we adopt a representation invariant to such choices by treating each edge as a straight segment in \(\mathbb{R}^3\) and thickening it based on a pseudo radius \(r>0\). We then define a continuous graph occupancy field \(\phi_r\), with a pseudo radius \(r\) assigned to each of its edges. To obtain the graph occupancy field at a point \(p\in\mathbb{R}^3\), we compute the distance \(d_G(p,G)\) to the closest edge (and, if present, isolated nodes) as follows:
\[
d_G(p,G)=\min\Bigg\{
\min_{(i,j)\in E} \big\|\,p-\big(p_i+\alpha_{ij}(p)\,(p_j-p_i)\big)\,\big\|,
\ \min_{i\in V}\|p-p_i\|
\Bigg\},
\]
\noindent where \(p_i\), \(p_j\) are node coordinates associated with edge \((i,j)\in E\), and
\[
\alpha_{ij}(p)=\operatorname{clamp}\!\left(\frac{(p-p_i)\cdot(p_j-p_i)}{\|p_j-p_i\|^2},\,0,\,1\right),
~~
\operatorname{clamp}(t,0,1)=\min\{1,\max\{0,t\}\}.
\]
\noindent The clamping ensures the closest point lies on the \emph{segment} and $\min_{i\in V}\|p-p_i\|$ is used to handle boundary nodes. From this, we obtain the graph occupancy field \(\phi_r\) as follows:
\begin{equation}
\phi_r(p,G) =
\begin{cases}
1, & \text{if } d_G(p,G) \le r\\[4pt]
0, & \text{otherwise}
\end{cases}.
\end{equation}
\noindent Note that the choice of the pseudo-radius \(r\) is critical as it determines how faithfully the occupancy field represents the graph topology. Too small \(r\) could hinder the learning capabilities of the model, whereas larger \(r\) could miss topologically important structures and thus, introduce artifacts. We provide a detailed ablation analysis of the pseudo radius choice in our experiments (see Section \ref{sec:rad_ablation}).

\subsection{VesselTok}
We present VesselTok, a method to generate latent representations arising from graph occupancy fields \(\phi_r\) conditioned on centerline graphs \(G\).
Prior methods developed for generic 3D shapes~\cite{zhang20233dshape2vecset,zhao2025hunyuan3d} primarily rely on surface point samples, which we argue are suboptimal for high surface-to-volume structures such as vascular graphs under realistic computational constraints. In contrast, VesselTok operates on centerline points, a substantially more computationally tractable representation than dense surface meshes, and leverages them as structural primitives to model the graph occupancy field directly. This design allows VesselTok to jointly process all centerline points without subsampling, efficiently producing expressive latent representations even for large and complex structures.
We adopt a VAE-style encoder-decoder architecture to obtain a continuous token $\mathbf{Z}$ for the occupancy field $\phi_r$, where the encoder $\mathcal{T}$ and decoder $\mathcal{D}$ consist of transformer blocks following recent works \cite{zhao2023michelangelo,zhao2025hunyuan3d}. 
An architectural overview of VesselTok is presented in Fig.~\ref{fig:arch}.

\paragraph{Encoder:} The encoder $\mathcal{T}$ aims to extract features from the centerline point cloud $\mathbf{P}$ and subsequently convert them into continuous tokens $\mathbf{Z}$. For this purpose, we initialize $L$ query points $\mathbf{Q}_{\text{in}}$ by using farthest point sampling (FPS) from $\mathbf{P}$. This results in an initial representation of the underlying structure, which is refined by the encoder. We use Fourier positional encoding for both the input point cloud $\mathbf{P}$ and the query points $\mathbf{Q}_{\text{in}}$, followed by a linear embedding layer to expand the feature dimension to the hidden dimension $d$. In the first layer of the encoder, we contextualize the input point cloud via the query points by applying a cross-attention layer between the embeddings of $\mathbf{P}$ and $\mathbf{Q}_{\text{in}}$. Next, we employ a series of self-attention layers to obtain the encoder hidden representation $\mathbf{H}\in \mathbb{R}^{l\times d}$. We obtain the mean and variance denoted by $\mathbf{Z}_{\mu},\mathbf{Z}_{\sigma}\in \mathbb{R}^{l\times c}$, respectively, via two final linear layers.

\paragraph{Decoder:} The decoder $\mathcal{D}$ reconstructs the graph occupancy field from the encoded latent. The first decoder layer maps the latent to the hidden dimension. Subsequently, the embedding goes through a series of self-attention layers. Next, we sample query points $\mathbf{Q}_{\text{out}}$ from the 3D grid, and embed them using Fourier positional encoding followed by a linear projection. Finally, the embeddings of the $\mathbf{Q}_{\text{out}}$ are forwarded to a cross-attention layer to predict the graph occupancy field $\Tilde{\phi}_\theta$ parameterized by the model parameters $\theta$.

\paragraph{Training:} We apply a reconstruction loss on the decoder output and the KL-divergence at the latent to train our model. For the reconstruction loss, we compute the Binary Cross Entropy (BCE) between the reconstructed graph occupancy and the reference on the query points. The total training loss is formalized as follows:
\begin{equation}
\mathcal{L}_{\text{total}} =\mathbb{E}_{p \in \mathbb{R}^3}[\operatorname{BCE}(\Tilde{\phi}_\theta(p), \phi(p,G))] + \lambda \cdot \operatorname{KL}(\mathcal{N}(\mathbf{Z}_{\mu},\operatorname{diag}(\mathbf{Z}_{\sigma})),\,\mathcal{N}(\mathbf{0},\mathbf{I}))\label{eqn:loss}.
\end{equation}

\noindent Since the graph occupancy field is quite sparse throughout the entire domain, we employ an imbalance-aware query selection strategy similar to \cite{zhang20233dshape2vecset}, emphasizing points located on the boundary of the object.

\paragraph{Inference:} \label{sec:graph_extraction} For graph reconstruction, we decode the latent representation into a continuous occupancy field and then recover a discrete centerline graph. We evaluate the predicted field on a regular grid \(\mathcal{X}\subset\mathbb{R}^3\), apply a threshold \(\tau\in(0,1)\), and obtain a discretized occupancy field:
\[
\Omega_\tau \;=\; \{\, p\in\mathcal{X}\;:\;\tilde{\phi}_\theta(p) \ge \tau \,\}.
\]
Next, we extract centerline points from the discretized occupancy field using a skeletonization algorithm~\cite{lee1994building}, yielding a set of predicted skeleton points \(\hat{V}\). We convert \(\hat{V}\) to a graph by a deterministic neighborhood-based connectivity prediction ($\hat{E}$).

This procedure yields a discrete, topology-preserving graph \(\hat{G}\) consistent with the continuous graph occupancy predicted by the decoder, closing the loop from latent decoding back to a usable centerline representation. We apply the same graph-extraction procedure to both our method and all baselines for a fair comparison. Importantly, this step is modular, and more robust graph extraction methods (e.g.,~\cite{meyer2009voreen}) can be substituted without changing the rest of the pipeline.

\section{Experiments}
\label{sec:exp}

\paragraph{Datasets:}
Similar to recent works~\cite{wittmann2025vesselfm}, we curate an expressive publicly available dataset spanning diverse anatomies (see supplementary Sec.~A). To this end, we extract graphs from airway datasets (ATM~\cite{zhang2023multi}, AIIB~\cite{nan2024hunting}, AeroPath~\cite{stoverud2023aeropath}), cerebral vasculature (COSTA~\cite{mou2024costa}), and pulmonary vessels (HiPas~\cite{chu2025deep}, PARSE~\cite{luo2023efficient}, Pulmonary-AV~\cite{cheng2024fusion}). For out-of-distribution validation, we evaluate our method on TopCoW~\cite{yang2025benchmarking} and renal vasculature data~\cite{walsh2021imaging,yagis2023deep,kuo2023terabyte}. Graphs are derived from segmentation masks using the Voreen graph extraction tool~\cite{meyer2009voreen}, yielding centerline graphs with per-edge geometry. The resulting dataset spans a wide range of sizes and topologies (\eg, node counts, branching factors, and loop prevalence). Importantly, all graphs are derived from real biomedical images, and no additional post-processing or refinement steps that could potentially perturb the graph structure were applied.

\paragraph{Metrics:}

To assess reconstruction quality, we compare the topology of reconstructed graphs to the ground truth using Betti numbers. Specifically, for topology, we report absolute Betti differences, $|\Delta \beta_0|$ (connected components) and $|\Delta \beta_1|$ (loops), where lower values indicate better preservation of graph structure. However, Betti-based errors alone are insufficient since, although they capture homology equivalence, they do not measure spatial coverage or geometric overlap between structures. Therefore, we complement them with clDice~\cite{shit2021cldice} and Chamfer Distance (CD). clDice emphasizes centerline fidelity and connectivity by quantifying overlap between the reconstructed and reference centerlines, whereas CD measures geometric discrepancy between the reconstructed and reference shapes. For all evaluations, we render graphs into a continuous occupancy field by dilating edges with a pseudo-radius of $r$ (see Section \ref{sec:rad_ablation}) on a 512\textsuperscript{3} voxel grid, and compute clDice and CD between the reconstruction and the corresponding ground-truth occupancy fields.

For generative evaluation, we report both point-based and graph-based metrics. We follow Zhang et al.~\cite{zhang20233dshape2vecset} and report Fréchet Inception Distance (FID), Maximum Mean Discrepancy based on Chamfer Distance (MMD-CD), and Earth Mover Distance (MMD-EMD) along with Coverage (COV-CD, COV-EMD). To compute FID, we train a PointNet++ encoder~\cite{qi2017pointnet++} to predict the anatomical labels of the centerlines contained in the training set.
The FID is computed between the embeddings of the training set samples and 1,000 generated samples. We also compute MMD and Coverage on Betti summaries (MMD–Betti, COV–Betti) of graphs to assess distributional alignment in terms of connected components and loops. For further details, please see supplementary Sec.~C.  The source code is publicly available at {\href{https://github.com/chinmay5/vessel_tok}{https://github.com/chinmay5/vessel\_tok}}

\begin{table}[t]
\centering
\scriptsize
\caption{
Quantitative results achieved on our testing anatomies. We report metrics for each dataset.
}
\label{tab:quantitative_results}
\begin{tabular}{c|l|c c c c}
\toprule
\quad\quad Data \quad\quad & Method & \;$\text{clDice}\uparrow$\; & \;$\text{CD}\downarrow$\; & \;$|\Delta \beta_0|\downarrow$\; & \;$|\Delta \beta_1|\downarrow$\;\\
\midrule
    \multirow{3}{*}{\rotatebox[origin=c]{0}{ATM}}
    & Hunyuan3D 2.0~\cite{zhao2025hunyuan3d} & 86.75 & 4.08 & 7.28 &  6.40\\
    & 3DShape2VecSet~\cite{zhang20233dshape2vecset} & 90.08 & 2.14 & 7.46 &  6.53\\
    & VesselTok (ours) & \textbf{96.33} & \textbf{0.96} & \textbf{4.91} &  \textbf{6.19}\\
\midrule
    \multirow{3}{*}{\rotatebox[origin=c]{0}{AIIB}}
    & Hunyuan3D 2.0~\cite{zhao2025hunyuan3d} & 85.13 & 2.90 & 8.40 & 11.00\\
    & 3DShape2VecSet~\cite{zhang20233dshape2vecset} & 87.22 & 2.74 & 8.6 & \textbf{10.85}\\
    & VesselTok (ours) & \textbf{94.85} & \textbf{0.66} & \textbf{4.47} & 11.00\\
\midrule
    \multirow{3}{*}{\rotatebox[origin=c]{0}{COSTA}}
    & Hunyuan3D 2.0~\cite{zhao2025hunyuan3d}  & 56.53 & 4.28 & 79.15 & 56.19\\
    & 3DShape2VecSet~\cite{zhang20233dshape2vecset} & 59.65 & 2.29 & 70.85 & 54.35 \\
    & VesselTok (ours) & \textbf{77.26} & \textbf{1.64} & \textbf{26.81} & \textbf{26.83}\\
\midrule
    \multirow{3}{*}{\rotatebox[origin=c]{0}{HiPas}}
    & Hunyuan3D 2.0~\cite{zhao2025hunyuan3d}  & 52.58 & 3.95 & 137.31 & \textbf{23.26}\\
    & 3DShape2VecSet~\cite{zhang20233dshape2vecset} & 55.58 & 2.92 & 132.56 & 28.65 \\
    & VesselTok (ours) & \textbf{67.59} & \textbf{1.82} & \textbf{115.88} & 38.42\\ 
\midrule
    \multirow{3}{*}{\rotatebox[origin=c]{0}{PARSE}}
    & Hunyuan3D 2.0~\cite{zhao2025hunyuan3d}  & 42.64 & 3.42 & 159.06 & 37.06 \\
    & 3DShape2VecSet~\cite{zhang20233dshape2vecset} & 42.44 & 3.19 & 155.06 & \textbf{32.31} \\
    & VesselTok (ours) & \textbf{57.03} & \textbf{0.69} & \textbf{128.06} & 34.06 \\
\midrule
    \multirow{3}{*}{\rotatebox[origin=c]{0}{Pulm.}}
    & Hunyuan3D 2.0~\cite{zhao2025hunyuan3d}  & 79.46 & 6.99 &  31.26  & 5.61 \\
    & 3DShape2VecSet~\cite{zhang20233dshape2vecset} & 86.40 & 1.59 & 25.52 & \textbf{5.04} \\
    & VesselTok (ours) & \textbf{94.30} & \textbf{0.74} &  \textbf{9.35}  & 5.52 \\

\bottomrule
\end{tabular}
\end{table}

\begin{figure*}[t!]
    \centering
    \includegraphics[width=0.9\textwidth]{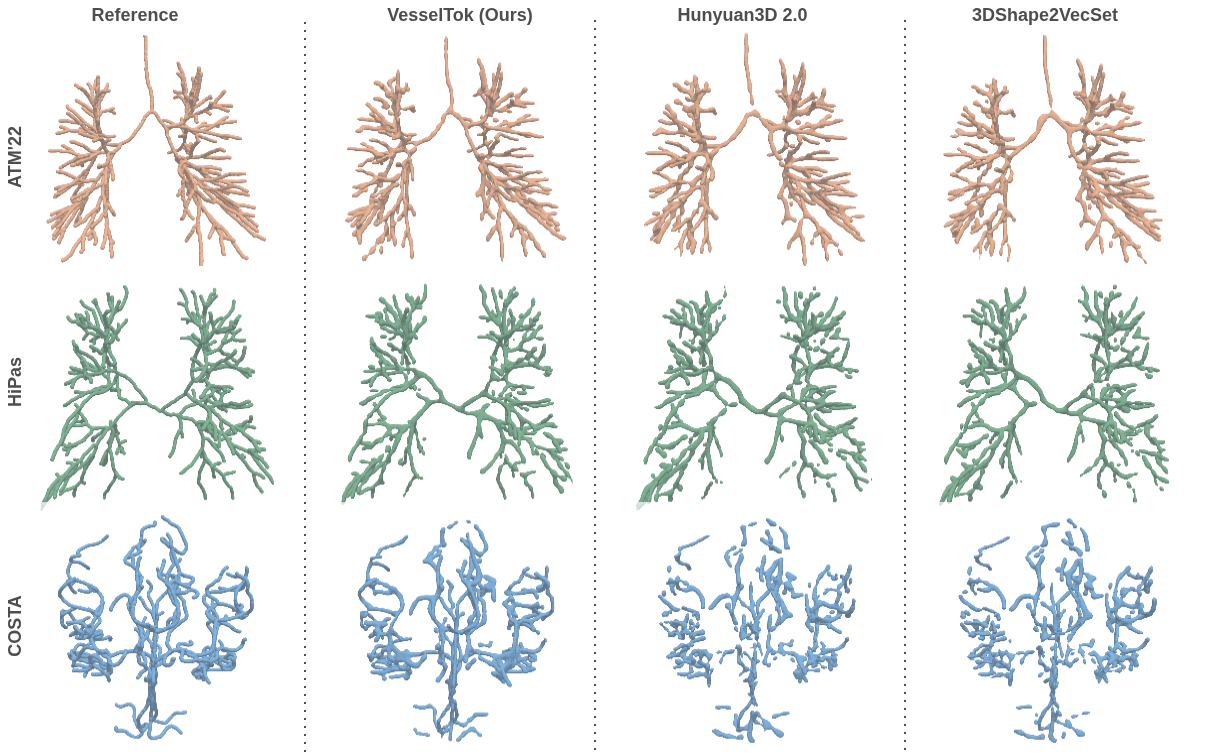}
    \caption{Qualitative results for the graph reconstruction task. We find that VesselTok demonstrates superior reconstruction capabilities.}
    \label{fig:qual}
\end{figure*}

\subsection{Reconstruction}
VesselTok delivers consistent gains over the strong baselines 3DShape2VecSet~\cite{zhang20233dshape2vecset} and Hunyuan3D 2.0~\cite{zhao2025hunyuan3d} across most datasets and metrics. The per-dataset train-test splits and implementation details are provided in the supplementary Sec.~A and D, respectively. Tab.~\ref{tab:quantitative_results} shows that VesselTok attains higher clDice, lower Chamfer Distance, and lower $|\Delta \beta_0|$ error consistently and competitive $|\Delta \beta_1|$ error. A per-dataset breakdown on airways (ATM and AIIB), cerebral vasculature (COSTA), and pulmonary vessels (HiPas, PARSE, and Pulmonary-AV) confirms that these improvements are consistent across anatomies. Notably, VesselTok better preserves connectivity and loop structure on average while reducing geometric discrepancy, indicating closer adherence to the underlying graph topology. Qualitatively, Fig.~\ref{fig:qual} shows superior reconstruction capabilities of our proposed method over the previous shape-encoding methods. Supplementary Sec.~E shows additional qualitative results.

\begin{table}[!t]
    \centering
    \scriptsize
    \caption{Comparison against vessel-specific and other baselines on the ATM dataset}
    \begin{tabular}{l|c c c c} 
    \toprule
         Method & \;$\text{clDice}\uparrow$\; & \;$\text{CD}\downarrow$\; & \;$|\Delta \beta_0|\downarrow$\; & \;$|\Delta \beta_1|\downarrow$\;\\
         \midrule
         VesselGPT~\cite{feldman2025vesselgpt} & 77.09 &  0.008 & \textbf{0.05} & 10.28 \\
         Hunyuan3D 2.0~\cite{zhao2025hunyuan3d}  & 87.31 & 0.007 & 0.11 & 9.31\\
         3DShape2VecSet~\cite{zhang20233dshape2vecset} & 90.11 & 0.007 & 0.09 & 9.48 \\
         VesselTok (ours) & \textbf{96.61} & \textbf{0.005} & 0.07 &  \textbf{8.97} \\
         \bottomrule
    \end{tabular}
    \label{tab:atm_comparison}
\end{table}

\subsection{Vessel Specific Baselines}
The state-of-the-art vessel-generation approach~\cite{prabhakar20243d} can handle branches and loops but operates in the uncompressed point cloud space, which makes it computationally intractable for the number of nodes in our dataset. Alternatively, we benchmark the reconstruction capabilities of VesselTok against established autoencoder VesselGPT~\cite{feldman2025vesselgpt}.
However, VesselGPT is a vessel-tree-only method. 

Hence, we compare VesselTok and other baselines only on the airway tree (ATM) dataset, training them from scratch.
Please note that airway trees are expected to be loop-free but often contain small spurious cycles due to minute inaccuracies in segmentation. Since VesselGPT assumes a strict tree structure, we needed to remove these loops during preprocessing. In contrast, VesselTok requires no such cleanup.
We also experimented with VesselVAE~\cite{feldman2023vesselvae}, but it fails to converge.
Tab.~\ref{tab:atm_comparison} shows that VesselTok achieves the best clDice, CD, and $|\Delta \beta_1|$, reflecting higher geometric fidelity and better recovery of cyclic structure, while VesselGPT performs best on $|\Delta \beta_0|$, likely benefiting from the simplification step in its preprocessing pipeline.

\begin{table}[t]
\centering
\scriptsize
\caption{
Quantitative results achieved on unseen anatomies and scales. 
}
\label{tab:quantitative_results_ood}
\begin{tabular}{c|l|c c c c}
\toprule
Data & Method & \;$\text{clDice}\uparrow$\; & \;$\text{CD}\downarrow $\; & \;$|\Delta \beta_0|\downarrow$\; & \;$|\Delta \beta_1|\downarrow$\;\\
\midrule
    \multirow{3}{*}{\rotatebox[origin=c]{0}{ATM$^{*}$}}
    & Hunyuan3D 2.0~\cite{zhao2025hunyuan3d} & 89.81 & 0.125 & 2.66 & 5.34\\
    & 3DShape2VecSet~\cite{zhang20233dshape2vecset} & 96.26 & 0.097 & \textbf{0.049} & 5.34\\
    & VesselTok (ours) & \textbf{99.16} & \textbf{0.001} & \textbf{0.049} & \textbf{5.29}\\
\midrule
    \multirow{3}{*}{\rotatebox[origin=c]{0}{COSTA$^{*}$}}
    & Hunyuan3D 2.0~\cite{zhao2025hunyuan3d}  & 81.16 & \textbf{0.121} & 0.85 & 5.20 \\
    & 3DShape2VecSet~\cite{zhang20233dshape2vecset} & 82.65 & 0.123 & 0.59 & 4.36 \\
    & VesselTok (ours) & \textbf{95.44} & 0.125 & \textbf{0.24} & \textbf{3.49} \\
\midrule
    \multirow{3}{*}{\rotatebox[origin=c]{0}{TopCoW}}
    & Hunyuan3D 2.0~\cite{zhao2025hunyuan3d}  & 97.82 & 0.003 & 0.09 & 0.21\\
    & 3DShape2VecSet~\cite{zhang20233dshape2vecset} & 97.96 & \textbf{0.001} & \textbf{0.06} & 0.14 \\
    & VesselTok (ours) & \textbf{99.42} & 0.002 & 0.07 & \textbf{0.04}\\
\midrule
    \multirow{3}{*}{\rotatebox[origin=c]{0}{RV}}
    & Hunyuan3D 2.0~\cite{zhao2025hunyuan3d}  & 73.81 & 0.086 & 15.29 & 6.68\\
    & 3DShape2VecSet~\cite{zhang20233dshape2vecset} & 79.66 & 0.019 & \textbf{11.91} & 6.45 \\
    & VesselTok (ours) & \textbf{88.86} & \textbf{0.017} & 13.09 & \textbf{6.25}\\ 
\bottomrule
\end{tabular}
\\
{\scriptsize\hfill\raggedright $^{*}$ Synthetic set created by partitioning vessels along the sagittal plane.\hfill}
\end{table}

\begin{figure}[t!]
    \centering
    \includegraphics[width=0.8\textwidth]{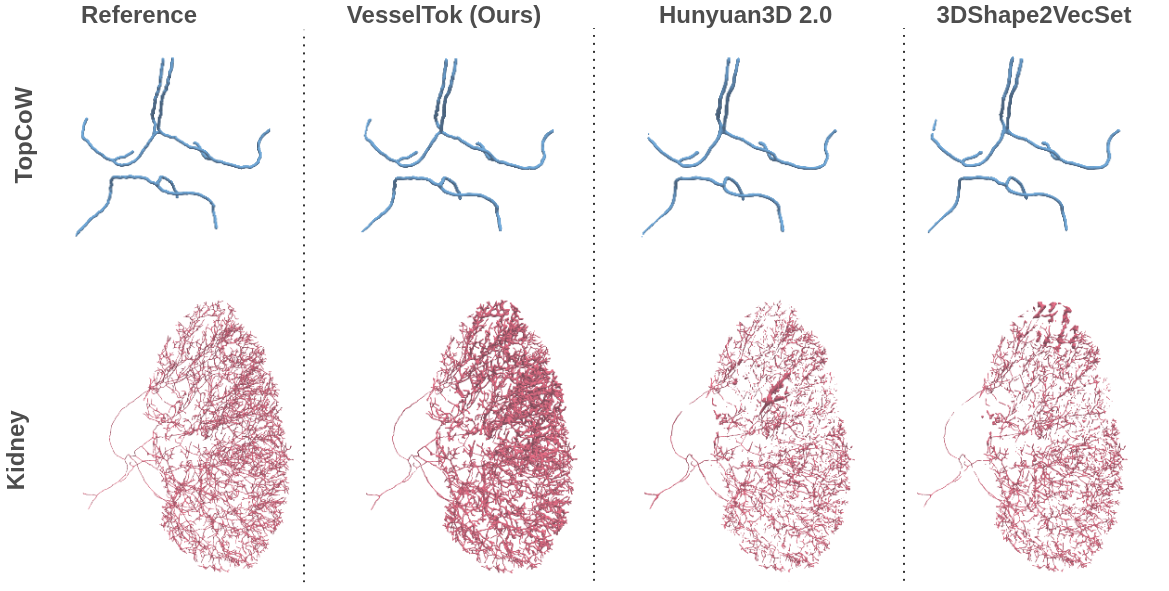}
    \caption{Qualitative results for the graph reconstruction task of previously unseen domains. This demonstrates VesselTok's strong prior, resulting in robust reconstructions.}
    \label{fig:ood}
\end{figure}

\subsection{Generalization to Unseen Anatomies} 

To assess VesselTok’s generalization, we evaluate it on anatomies and graph scales outside the training distribution. VesselTok is trained on airways, whole-brain vessels, and pulmonary trees, with graphs ranging from $\approx$2500 to $>$ 10000 nodes. To demonstrate its generalizability, we applied VesselTok on three datasets: (i) large-scale renal vasculature (RV), (ii) compact human Circle of Willis (CoW), and (iii) synthetic data created by partitioning airway trees and brain vessels along the sagittal plane to induce anatomically implausible edits (probing topological breaks and long-range consistency). CoW graphs are small (< 1,000 nodes on average), while renal graphs exceed 100,000 nodes (over an order of magnitude larger than typical training examples). Although not limited by computational scalability, to obtain higher fidelity, we ran inference via spatial chunking with $150^3$ grid size, yielding heterogeneous chunks ranging from sparse (< 200 nodes) to dense (> 8,000 nodes). For all graphs, coordinates are normalized to [-1,1] and centered by subtracting the center of mass.

Tab.~\ref{tab:quantitative_results_ood} and Fig.~\ref{fig:ood} show that VesselTok robustly encodes heterogeneous out-of-distribution structures, achieving a clDice of 99.42 on CoW and 88.86 on RV, while also attaining lower $\beta_1$ error and competitive $\beta_0$ error, indicating more faithful topology preservation than the baselines. On the synthetic sagittal-cut ATM and COSTA samples, VesselTok maintains clDice scores of 99.16 and 95.44, respectively, demonstrating resilience to anatomically implausible topological edits. In contrast, baseline methods consistently exhibit lower clDice and higher $\beta_1$ error (i.e., more spurious or missed loops), while 3DShape2VecSet achieves competitive $\beta_0$ error. For additional details, see supplementary Sec~F.

\begin{table*}[t]
\centering
\scriptsize
\caption{FID computed in the PointNet++ latent space, Maximum Mean Discrepancy on Chamfer and Earth Mover distance, and Betti error for the generated samples using EDM~\cite{karras2022elucidating} on the latent tokens. We present conditional and unconditional results. MMD-CD and MMD-EMD values are reported in 10$^{-2}$.}
\begin{tabular}{c l | c  c c c c  c c c c}
\toprule
& \multirow{2}{*}{Method} & \multirow{2}{*}{FID$\;\downarrow$} &
\multicolumn{4}{c}{MMD$\;\downarrow$} &
\multicolumn{4}{c}{COV$\;\uparrow$} \\
\cmidrule(l{2pt}r{2pt}){4-7} \cmidrule(l{2pt}r{2pt}){8-11}
&&& \;CD\; & \;EMD\; & \;$\beta_0$\; & \;$\beta_1$\; & \;CD\; & \;EMD\; & \;$\beta_0$\; & \;$\beta_1$\; \\
\midrule
\multirow{3}{*}{\rotatebox[origin=c]{90}{cond.}\quad}
& Hunyuan3D 2.0~\cite{zhao2025hunyuan3d}            & 74.61 & 1.89 & 0.17 & 28.06 & 5.19 & 0.43 & 0.46 & 0.84 & 0.68 \\
& 3DShape2VecSet~\cite{zhang20233dshape2vecset}     & 73.58 & 1.92 & 0.18 & 24.07 & 6.82 & 0.44 & 0.50 & 0.89 & 0.61 \\
& VesselTok (ours)                                  & \textbf{43.13} & \textbf{0.25} & \textbf{0.14} & \textbf{7.25} & \textbf{1.01} & \textbf{0.48} & \textbf{0.53} & \textbf{0.95} & \textbf{0.73} \\
\midrule
\multirow{3}{*}{\rotatebox[origin=c]{90}{uncon.}\quad} 
& Hunyuan3D 2.0~\cite{zhao2025hunyuan3d}            & 107.57 & 1.85 & 3.51 & 125.68 & 3.74 & 0.40 & 0.28 & 0.22 & 0.12 \\
& 3DShape2VecSet~\cite{zhang20233dshape2vecset}     & 146.57 & 1.88 & 3.45 & 127.60 & 3.22 & 0.39 & 0.26 & 0.21 & 0.08 \\
& VesselTok (ours)                                  & \textbf{96.87} & \textbf{0.30} & \textbf{2.35} & \textbf{78.36} & \textbf{2.64} & \textbf{0.42} & \textbf{0.32} & \textbf{0.32} & \textbf{0.18} \\
\bottomrule
\end{tabular}
\label{tab:FID_betti}
\end{table*}

\begin{figure*}[t!]
    \centering
    \includegraphics[width=\textwidth]{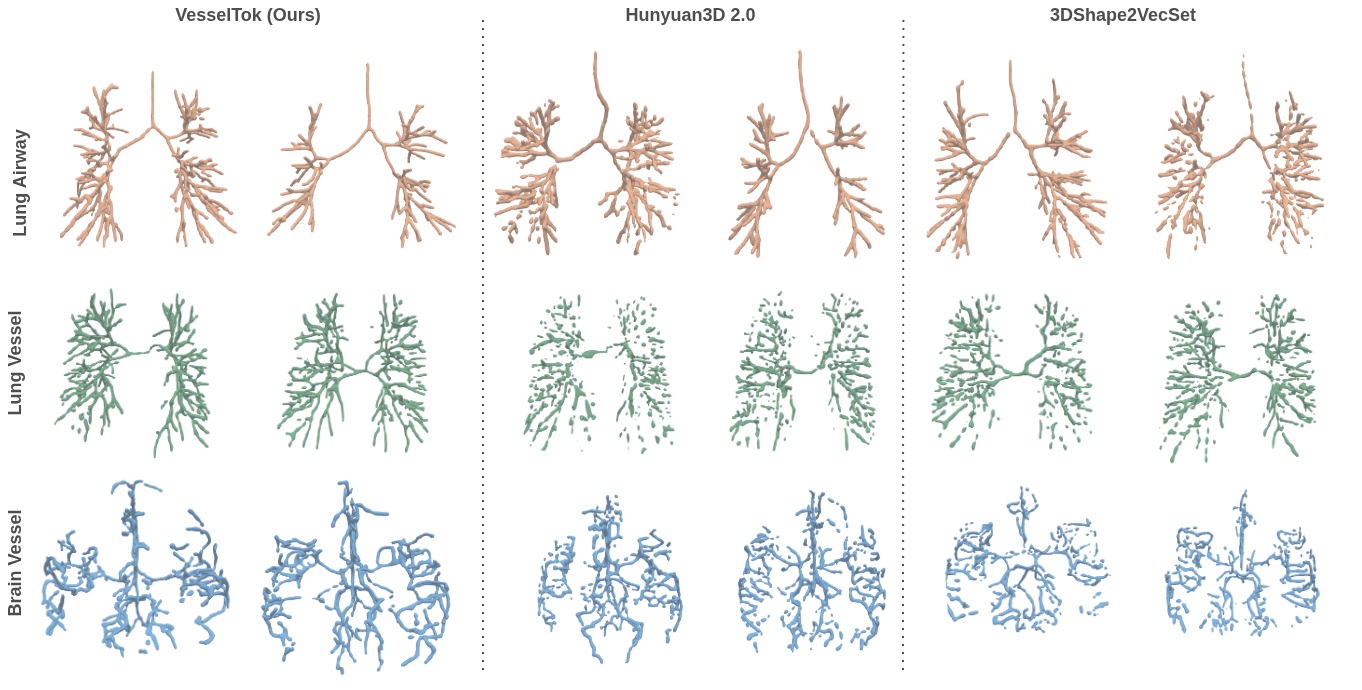}
    \caption{Qualitative results for conditional generation. VesselTok consistently generates more realistic vessels than previous methods.}
    \label{fig:cond}
\end{figure*} 

\subsection{Generative Modeling}

We generate anatomical graphs by training a diffusion model in the token space. Using a fixed-length sequence of 512 tokens per graph keeps training and sampling tractable compared to native graph generation \cite{prabhakar20243d}. We train both unconditional and class-conditional models (conditioned on anatomical categories), using the EDM backbone \cite{karras2022elucidating} and hyperparameters matching 3DShape2VecSet~\cite{zhang20233dshape2vecset}. We compare our method against the strong baselines Hunyuan3D~\cite{zhao2025hunyuan3d} and 3DShape2VecSet, while keeping the architecture and training budget fixed to ensure fairness. Please see Supplementary Sec. G for details.

\begin{figure}[!t]
\centering
\begin{minipage}[t]{0.48\textwidth}
  \centering
  \includegraphics[width=\textwidth,trim=8mm 8mm 8mm 0mm, clip]{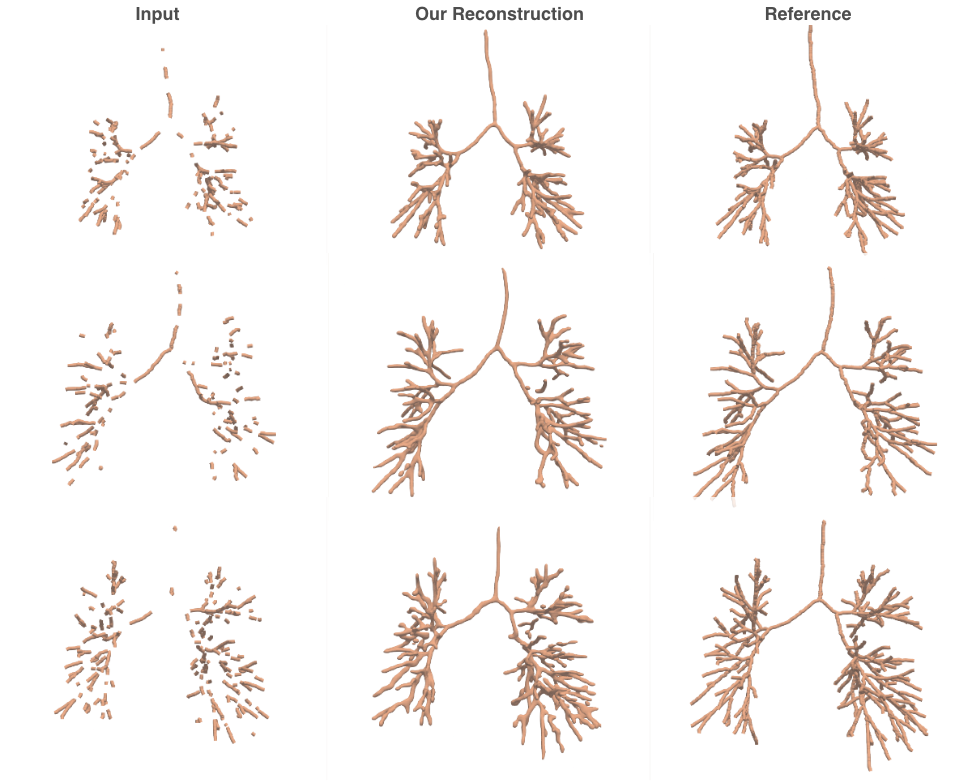}
  \captionof{figure}{Qualitative examples for link prediction on the airway tree (ATM) dataset. Given an incomplete airway graph with missing connections, our model infers and inserts plausible edges to repair the centerline while preserving anatomical plausibility.}
  \label{fig:link_pred_main}
\end{minipage}\hfill
\begin{minipage}[t]{0.48\textwidth}
  \centering
  \includegraphics[width=\textwidth]{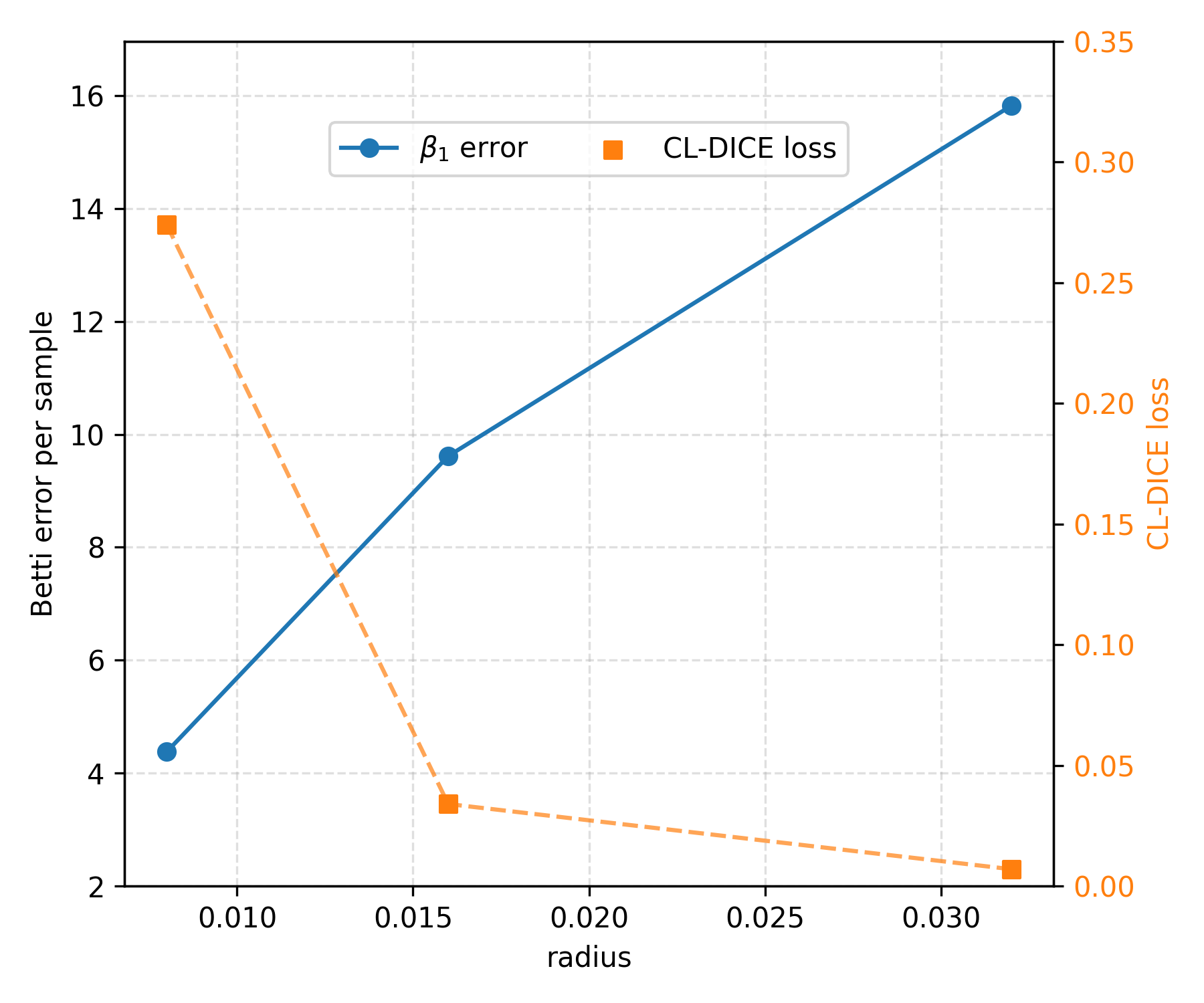}
  \captionof{figure}{Comparison of reconstruction fidelity vs.\ topological characteristics retained as we change the pseudo radius $r$. Lower values of $r$ improve topological characteristics ($|\Delta \beta_1|$) but degrade reconstruction performance (clDice loss). We find $r=0.016$ strikes a good balance.}
  \label{fig:pseudo_radius}
\end{minipage}
\end{figure}

As reported in Table \ref{tab:FID_betti} and Fig.~\ref{fig:cond}, our latent yields lower FID in both unconditional and conditional regimes and also improves complementary metrics (MMD-CD, MMD-EMD, COV-CD, and COV-EMD). Further, the lower Betti errors indicate that graphs generated by our method align with the ground truth data distribution better. This demonstrates that a semantically structured token space materially benefits generative modeling of biomedical, spatial graphs. Additional training and ablation details are mentioned in supplementary Sec. D.\\

\begin{table}[!b]
\centering
\scriptsize
\caption{VesselTok's performance on inverse problems. On the ATM infill task, VesselTok outperforms the baseline ~\cite{amiranashvili2024learning}, producing more coherent vascular structures.} 
\begin{tabular}{l|c c c c} 
\toprule 
Method & clDice $\uparrow$ & CD$\downarrow$ & $|\Delta \beta_0|\downarrow$ & $|\Delta \beta_1|\downarrow$ \\ 
\midrule Autodecoder~\cite{amiranashvili2024learning} & 83.49 & 0.066 & 4.32 & 8.68 \\ 
VesselTok (ours) & \textbf{88.13} & \textbf{0.043} & \textbf{3.19} & \textbf{8.58} \\ 
\bottomrule \end{tabular}
\label{tab:inpainting} 
\end{table}

\subsection{Downstream Tasks}
Beyond unconditional generation, the learned token representation supports clinically relevant downstream tasks such as link prediction for repairing incomplete vasculature. Specifically, given a vessel graph with missing connections, the goal is to infer plausible edges that are consistent with both local geometry and global topology. For this, we train a Diffusion Transformer~\cite{peebles2023scalable} with a conditional flow-matching objective~\cite{lipman2022flow}. The incomplete graph serves as a conditioning signal, and the model learns to denoise toward the distribution of the complete (clean) graph. Training pairs are generated by the description provided in supplementary Sec.~H, resulting in a graph with $\approx$40\% of missing links. The model learns to recover the missing connectivity. In inference, the model samples plausible completions that "infill" the missing edges. Tab.~\ref{tab:inpainting} shows that our approach performs competitively with a strong autodecoder baseline~\cite{amiranashvili2024learning}. Fig.~\ref{fig:link_pred_main} shows some qualitative examples.

\subsection{Ablation Studies}
We ablate two core design choices of VesselTok: (i) the pseudo-radius $r$ used to form the occupancy field, and (ii) the number $K$ and channel dimension $C$ of the tokens in the VAE latent space.

\paragraph{Pseudo Radius:} \label{sec:rad_ablation}
We model each biomedical graph as a 3D centerline. Conventional neural fields struggle with extremely thin structures~\cite{sitzmann2020implicit}, while vector-field–based approaches for curve-like geometry have shown limited fidelity~\cite{mello2025neural}. In contrast, our objective prioritizes topological connectivity over recovering physical thickness. Hence, we adopt a simple design choice: convert the graph to a continuous occupancy field by dilating each edge with a fixed pseudo-radius $r$. Please note that using this continuous representation decouples synthesis from an a priori node count, an assumption required by several existing methods~\cite{vignac2023midi, prabhakar20243d}.

To study the trade-off between reconstruction fidelity and topology preservation as a function of the dilation radius $r$, we train three encoders on ATM with $r \in {0.008, 0.016, 0.032}$. Fig.~\ref{fig:pseudo_radius} shows that larger $r$ simplifies learning and improves reconstruction (clDice loss $\approx 0.007$) but degrades fine topology ($\beta_1$ error $\approx 16$ per sample), whereas smaller $r$ better preserves topology ($\beta_1$ error $\approx 4$) but reduces reconstruction quality (clDice loss $\approx 0.28$). Hence, we select $r$ = 0.016 as a balanced operating point, which maintains salient topological features while achieving competitive reconstruction quality. Importantly, this pseudo-radius is used as a single global hyperparameter across experiments, rather than being tuned separately for each dataset.

\begin{table*}[b]
\centering
\scriptsize

\begin{minipage}[t]{0.49\textwidth}
\centering
\caption{Effect of varying the number of keypoints $K$ on reconstruction quality. Performance drops sharply at 256 or fewer keypoints, while 512 provides a sweet spot between reconstruction quality and compression ratio. ($C=4$ for all models).}
\begin{tabular}{c|c c c c c}
\toprule
$K$ & clDice $\uparrow$ & CD $\downarrow$ &  $|\Delta \beta_0|\downarrow$ & $|\Delta \beta_1|\downarrow$ & $\kappa$\\
\midrule
768   & \textbf{97.13} & \textbf{0.005}  & \textbf{0.07} & \textbf{8.93} & 4.68 \\ 
512   & 96.61 & \textbf{0.005}  & \textbf{0.07} & 8.97 & 7.03 \\ 
256   & 80.35 & 0.006 & 25.61 & 10.02 & 14.06\\
128   & 12.29 & 0.114  & 1.90 & 10.39 & 28.12\\ 
64    & 8.42 & 0.116  & 2.52 & 10.27 & 56.24\\ 
\bottomrule
\end{tabular}
\label{tab:keypoints}
\end{minipage}\hfill
\begin{minipage}[t]{0.48\textwidth}
\centering
\caption{Effect of channel dimension $C$ on reconstruction quality. Increasing the channel dimension improves the reconstruction, but at the expense of the compression ratio $\kappa$. We fix $K$ to 512 for this ablation.}
\begin{tabular}{c|c c c c c }
\toprule
$C$ & clDice $\uparrow$ & CD $\downarrow$ & $|\Delta \beta_0|\downarrow$ & $|\Delta \beta_1|\downarrow$ &  $\kappa$\\
\midrule
16 & 96.32 & \textbf{0.004} & \textbf{0.06} & \textbf{8.52} & 1.76\\
8  & 95.95 & \textbf{0.004} & \textbf{0.06} & 8.80 & 3.51 \\
4  & \textbf{96.61} & 0.005 & 0.07 & 8.97 & 7.03\\
2  & 7.08 & 0.058 & 1.68 & 10.36 & 14.06 \\
\bottomrule
\end{tabular}
\label{tab:channel_wise}
\end{minipage}

\end{table*}

\paragraph{VAE Latent Space:}
The number of tokens per graph, $K$, and the latent channel dimension, $C$, govern a tradeoff between compression and reconstruction fidelity. Since biomedical graphs exhibit substantial variation across individual anatomies, we use the average compression ratio, $\kappa$, to quantify this tradeoff. For a dataset consisting of $M$ samples, with each sample consisting of $N_i$ nodes, $\kappa$ is defined as:
$
    \kappa = \frac{1}{M} \sum_{i=1}^{M} \frac{N_i \cdot 3}{K \cdot C}
$
Our goal is to learn a \emph{compact} latent space that can faithfully reconstruct the graphs. We ablate both factors $K$ and $C$ by training VAEs from scratch on the representative ATM dataset. Reconstruction quality is measured with clDice scores, CD, and Betti error.
 
Tab.~\ref{tab:keypoints} varies $K$ and demonstrates that increasing $K$ improves reconstruction quality (clDice 97.13 and CD 0.005), but reduces $\kappa$ as a tradeoff. Small $K$ sharply degrades reconstruction quality (\eg, clDice 8.42 and CD 0.116 at $K$ = 64). Tab.~\ref{tab:channel_wise} varies $C$ and exhibits the same trend: higher $C$ yields better fidelity (\eg, $C$ = 16 reaches clDice 96.32 and CD 0.004) at the expense of compression. We find that overall, $K$ = 512 and $C$ = 4 offer a balanced operating point, achieving clDice of 96.61, CD of 0.005, and low Betti error. We adopt this configuration for our experiments.

\paragraph{\textbf{Limitation}}
A natural limitation of fixed-capacity latent representations is that performance can degrade as input complexity increases. However, our analysis suggests that graph size alone is not the primary bottleneck. We stratify samples into medium ($\leq$5.5K nodes) and large ($> $5.5K nodes) groups, using node count as a coarse proxy for complexity. Under this stratification, clDice drops only modestly on ATM (98.90 $\rightarrow$ 91.43), but decreases sharply on the more topologically complex COSTA graphs (89.19 $\rightarrow$ 75.20). This suggests that reconstruction performance is influenced more strongly by topological complexity than by graph size alone.

\section{Conclusion}

Existing methods struggle to generate large graphs due to computational bottlenecks, rendering them unsuitable for analyzing network-like structures in biomedical systems (\eg, vascular networks or airways). To address this issue, we present VesselTok, a novel graph tokenization method designed to learn a semantically rich, compressed latent representation of large networks. Unlike shape tokenizers that operate on mesh surfaces, VesselTok uses a centerline-based representation (occupancy around dilated centerlines), which is especially well-suited to high surface-to-volume geometries, such as dense vasculature. We demonstrate that VesselTok generalizes to anatomies not encountered during training and can learn compact latent representations of diverse anatomical networks. These latents can be efficiently leveraged for both generative modeling (unconditional and conditional synthesis) and inverse problem tasks (\eg, link prediction), where VesselTok consistently outperforms the state of the art, paving the way for large-scale, targeted analysis of biomedical networks.

\section*{Acknowledgments} This work has been supported by the Helmut Horten Foundation. S. Shit is supported by the UZH Postdoc Grant (K-74851-03-01).
 


%
%
\bibliographystyle{splncs04}
\bibliography{main}

@String(CVPR= {IEEE Conf. Comput. Vis. Pattern Recog.})

@String(TOG= {ACM Trans. Graph.})

@String(CVPR  = {CVPR})

@String(TOG   = {ACM TOG})

@article{zhang20233dshape2vecset,
  title={{3DShape2VecSet: A 3D Shape Representation for Neural Fields and
Generative Diffusion Models}},
  author={Zhang, Biao and Tang, Jiapeng and Niessner, Matthias and Wonka, Peter},
  journal={ACM Transactions On Graphics (TOG)},
  volume={42},
  number={4},
  pages={1--16},
  year={2023},
  publisher={ACM New York, NY, USA}
}

@article{zhao2025hunyuan3d,
  title={Hunyuan3d 2.0: Scaling diffusion models for high resolution textured 3d assets generation},
  author={Zhao, Zibo and Lai, Zeqiang and Lin, Qingxiang and Zhao, Yunfei and Liu, Haolin and Yang, Shuhui and Feng, Yifei and Yang, Mingxin and Zhang, Sheng and Yang, Xianghui and others},
  journal={arXiv preprint arXiv:2501.12202},
  year={2025}
}

@article{karras2022elucidating,
  title={Elucidating the design space of diffusion-based generative models},
  author={Karras, Tero and Aittala, Miika and Aila, Timo and Laine, Samuli},
  journal={Advances in neural information processing systems},
  volume={35},
  pages={26565--26577},
  year={2022}
}

@article{qi2017pointnet++,
  title={Pointnet++: Deep hierarchical feature learning on point sets in a metric space},
  author={Qi, Charles Ruizhongtai and Yi, Li and Su, Hao and Guibas, Leonidas J},
  journal={Advances in neural information processing systems},
  volume={30},
  year={2017}
}

@article{lee1994building,
  title={Building skeleton models via 3-D medial surface axis thinning algorithms},
  author={Lee, Ta-Chih and Kashyap, Rangasami L and Chu, Chong-Nam},
  journal={CVGIP: graphical models and image processing},
  volume={56},
  number={6},
  pages={462--478},
  year={1994},
  publisher={Elsevier}
}

@article{mello2025neural,
  title={Neural vector fields for implicit surface representation and inference},
  author={Mello Rella, Edoardo and Chhatkuli, Ajad and Konukoglu, Ender and Van Gool, Luc},
  journal={International Journal of Computer Vision},
  volume={133},
  number={4},
  pages={1855--1878},
  year={2025},
  publisher={Springer}
}

@article{sitzmann2020implicit,
  title={Implicit neural representations with periodic activation functions},
  author={Sitzmann, Vincent and Martel, Julien and Bergman, Alexander and Lindell, David and Wetzstein, Gordon},
  journal={Advances in neural information processing systems},
  volume={33},
  pages={7462--7473},
  year={2020}
}

@inproceedings{feldman2023vesselvae,
  title={Vesselvae: Recursive variational autoencoders for 3d blood vessel synthesis},
  author={Feldman, Paula and Fainstein, Miguel and Siless, Viviana and Delrieux, Claudio and Iarussi, Emmanuel},
  booktitle={International Conference on Medical Image Computing and Computer-Assisted Intervention},
  pages={67--76},
  year={2023},
  organization={Springer}
}

@inproceedings{feldman2025vesselgpt,
  title={VesselGPT: Autoregressive Modeling of Vascular Geometry},
  author={Feldman, Paula and Sinnona, Martin and Delrieux, Claudio and Siless, Viviana and Iarussi, Emmanuel},
  booktitle={International Conference on Medical Image Computing and Computer-Assisted Intervention},
  pages={662--672},
  year={2025},
  organization={Springer}
}

@article{batten2025vector,
  title={Vector Representations of Vessel Trees},
  author={Batten, James and Schaap, Michiel and Sinclair, Matthew and Bai, Ying and Glocker, Ben},
  journal={arXiv preprint arXiv:2506.11163},
  year={2025}
}

@article{chang20243d,
  title={3D Shape Tokenization via Latent Flow Matching},
  author={Chang, Jen-Hao Rick and Wang, Yuyang and Martin, Miguel Angel Bautista and Gu, Jiatao and Zhao, Xiaoming and Susskind, Josh and Tuzel, Oncel},
  journal={arXiv preprint arXiv:2412.15618},
  year={2024}
}

@article{zhao2023michelangelo,
  title={Michelangelo: Conditional 3d shape generation based on shape-image-text aligned latent representation},
  author={Zhao, Zibo and Liu, Wen and Chen, Xin and Zeng, Xianfang and Wang, Rui and Cheng, Pei and Fu, Bin and Chen, Tao and Yu, Gang and Gao, Shenghua},
  journal={Advances in neural information processing systems},
  volume={36},
  pages={73969--73982},
  year={2023}
}

@inproceedings{xiang2025structured,
  title={Structured 3d latents for scalable and versatile 3d generation},
  author={Xiang, Jianfeng and Lv, Zelong and Xu, Sicheng and Deng, Yu and Wang, Ruicheng and Zhang, Bowen and Chen, Dong and Tong, Xin and Yang, Jiaolong},
  booktitle={Proceedings of the Computer Vision and Pattern Recognition Conference},
  pages={21469--21480},
  year={2025}
}

@article{jun2023shap,
  title={Shap-e: Generating conditional 3d implicit functions},
  author={Jun, Heewoo and Nichol, Alex},
  journal={arXiv preprint arXiv:2305.02463},
  year={2023}
}

@inproceedings{chen2025dora,
  title={Dora: Sampling and benchmarking for 3d shape variational auto-encoders},
  author={Chen, Rui and Zhang, Jianfeng and Liang, Yixun and Luo, Guan and Li, Weiyu and Liu, Jiarui and Li, Xiu and Long, Xiaoxiao and Feng, Jiashi and Tan, Ping},
  booktitle={Proceedings of the Computer Vision and Pattern Recognition Conference},
  pages={16251--16261},
  year={2025}
}

@article{bensadoun2024meta,
  title={Meta 3d gen},
  author={Bensadoun, Raphael and Monnier, Tom and Kleiman, Yanir and Kokkinos, Filippos and Siddiqui, Yawar and Kariya, Mahendra and Harosh, Omri and Shapovalov, Roman and Graham, Benjamin and Garreau, Emilien and others},
  journal={arXiv preprint arXiv:2407.02599},
  year={2024}
}

@article{zhang20243d,
  title={3D representation in 512-Byte: Variational tokenizer is the key for autoregressive 3D generation},
  author={Zhang, Jinzhi and Xiong, Feng and Xu, Mu},
  journal={arXiv preprint arXiv:2412.02202},
  year={2024}
}

@article{zhang2024g3pt,
  title={G3pt: Unleash the power of autoregressive modeling in 3d generation via cross-scale querying transformer},
  author={Zhang, Jinzhi and Xiong, Feng and Xu, Mu},
  journal={arXiv preprint arXiv:2409.06322},
  year={2024}
}

@inproceedings{zhang2024implicit,
  title={Implicit Representation Embraces Challenging Attributes of Pulmonary Airway Tree Structures},
  author={Zhang, Minghui and Zhang, Hanxiao and You, Xin and Yang, Guang-Zhong and Gu, Yun},
  booktitle={International Conference on Medical Image Computing and Computer-Assisted Intervention},
  pages={546--556},
  year={2024},
  organization={Springer}
}

@inproceedings{wittmann2025vesselfm,
  title={{vesselFM: A Foundation Model for Universal 3D Blood Vessel Segmentation}},
  author={Wittmann, Bastian and Wattenberg, Yannick and Amiranashvili, Tamaz and Shit, Suprosanna and Menze, Bjoern},
  booktitle={Proceedings of the Computer Vision and Pattern Recognition Conference},
  pages={20874--20884},
  year={2025}
}

@inproceedings{shit2021cldice,
  title={clDice-a novel topology-preserving loss function for tubular structure segmentation},
  author={Shit, Suprosanna and Paetzold, Johannes C and Sekuboyina, Anjany and Ezhov, Ivan and Unger, Alexander and Zhylka, Andrey and Pluim, Josien PW and Bauer, Ulrich and Menze, Bjoern H},
  booktitle={Proceedings of the IEEE/CVF conference on computer vision and pattern recognition},
  pages={16560--16569},
  year={2021}
}

@inproceedings{vignac2023midi,
  title={Midi: Mixed graph and 3d denoising diffusion for molecule generation},
  author={Vignac, Clement and Osman, Nagham and Toni, Laura and Frossard, Pascal},
  booktitle={Joint European Conference on Machine Learning and Knowledge Discovery in Databases},
  pages={560--576},
  year={2023},
  organization={Springer}
}

@inproceedings{prabhakar20243d,
  title={3d vessel graph generation using denoising diffusion},
  author={Prabhakar, Chinmay and Shit, Suprosanna and Musio, Fabio and Yang, Kaiyuan and Amiranashvili, Tamaz and Paetzold, Johannes C and Li, Hongwei Bran and Menze, Bjoern},
  booktitle={International Conference on Medical Image Computing and Computer-Assisted Intervention},
  pages={3--13},
  year={2024},
  organization={Springer}
}

@article{zhang2023multi,
  title={Multi-site, multi-domain airway tree modeling},
  author={Zhang, Minghui and Wu, Yangqian and Zhang, Hanxiao and Qin, Yulei and Zheng, Hao and Tang, Wen and Arnold, Corey and Pei, Chenhao and Yu, Pengxin and Nan, Yang and others},
  journal={Medical image analysis},
  volume={90},
  pages={102957},
  year={2023},
  publisher={Elsevier}
}

@article{mou2024costa,
  title={COSTA: A multi-center TOF-MRA dataset and a style self-consistency network for cerebrovascular segmentation},
  author={Mou, Lei and Lin, Jinghui and Zhao, Yifan and Liu, Yonghuai and Ma, Shaodong and Zhang, Jiong and Lv, Wenhao and Zhou, Tao and Liu, Jiang and Frangi, Alejandro F and others},
  journal={IEEE transactions on medical imaging},
  volume={43},
  number={12},
  pages={4442--4456},
  year={2024},
  publisher={IEEE}
}

@article{yang2025benchmarking,
  title={{Benchmarking the cow with the topcow challenge: Topology-aware anatomical segmentation of the circle of willis for CTA and MRA}},
  author={Yang, Kaiyuan and Musio, Fabio and Ma, Yihui and Juchler, Norman and Paetzold, Johannes C and Al-Maskari, Rami and H{\"o}her, Luciano and Li, Hongwei Bran and Hamamci, Ibrahim Ethem and Sekuboyina, Anjany and others},
  journal={arXiv preprint arXiv:2312.17670},
  year={2025}
}

@article{meyer2009voreen,
  title={Voreen: A rapid-prototyping environment for ray-casting-based volume visualizations},
  author={Meyer-Spradow, Jennis and Ropinski, Timo and Mensmann, J{\"o}rg and Hinrichs, Klaus},
  journal={IEEE Computer Graphics and Applications},
  volume={29},
  number={6},
  pages={6--13},
  year={2009},
  publisher={IEEE}
}

@article{wang2016skeleton,
  title={Skeleton-based cerebrovascular quantitative analysis},
  author={Wang, Xingce and Liu, Enhui and Wu, Zhongke and Zhai, Feifei and Zhu, Yi-Cheng and Shui, Wuyang and Zhou, Mingquan},
  journal={BMC medical imaging},
  volume={16},
  number={1},
  pages={68},
  year={2016},
  publisher={Springer}
}

@article{epp2023role,
  title={The role of leptomeningeal collaterals in redistributing blood flow during stroke},
  author={Epp, Robert and Gl{\"u}ck, Chaim and Binder, Nadine Felizitas and El Amki, Mohamad and Weber, Bruno and Wegener, Susanne and Jenny, Patrick and Schmid, Franca},
  journal={PLoS computational biology},
  volume={19},
  number={10},
  pages={e1011496},
  year={2023},
  publisher={Public Library of Science San Francisco, CA USA}
}

@article{lin2022incomplete,
  title={Incomplete circle of {W}illis variants and stroke outcome},
  author={Lin, Eaton and Kamel, Hooman and Gupta, Ajay and RoyChoudhury, Arindam and Girgis, Peter and Glodzik, Lidia},
  journal={European Journal of Radiology},
  volume={153},
  pages={110383},
  year={2022},
  publisher={Elsevier}
}

@article{ortiz2023morphometric,
  title={Morphometric analysis of airways in pre-COPD and mild {COPD} lungs using continuous surface representations of the bronchial lumen},
  author={Ortiz-Puerta, David and Diaz, Orlando and Retamal, Jaime and Hurtado, Daniel E},
  journal={Frontiers in Bioengineering and Biotechnology},
  volume={11},
  pages={1271760},
  year={2023},
  publisher={Frontiers Media SA}
}

@article{belchi2018lung,
  title={Lung topology characteristics in patients with chronic obstructive pulmonary disease},
  author={Belchi, Francisco and Pirashvili, Mariam and Conway, Joy and Bennett, Michael and Djukanovic, Ratko and Brodzki, Jacek},
  journal={Scientific reports},
  volume={8},
  number={1},
  pages={5341},
  year={2018},
  publisher={Nature Publishing Group UK London}
}

@article{dabbah2011automatic,
  title={Automatic analysis of diabetic peripheral neuropathy using multi-scale quantitative morphology of nerve fibres in corneal confocal microscopy imaging},
  author={Dabbah, Mohammad A and Graham, Jim and Petropoulos, Ioannis N and Tavakoli, Mitra and Malik, Rayaz A},
  journal={Medical image analysis},
  volume={15},
  number={5},
  pages={738--747},
  year={2011},
  publisher={Elsevier}
}

@article{chapman2015automated,
  title={Automated generation of directed graphs from vascular segmentations},
  author={Chapman, Brian E and Berty, Holly P and Schulthies, Stuart L},
  journal={Journal of biomedical informatics},
  volume={56},
  pages={395--405},
  year={2015},
  publisher={Elsevier}
}

@article{bumgarner2022open,
  title={Open-source analysis and visualization of segmented vasculature datasets with {V}essel{V}io},
  author={Bumgarner, Jacob R and Nelson, Randy J},
  journal={Cell reports methods},
  volume={2},
  number={4},
  year={2022},
  publisher={Elsevier}
}

@article{yang2023vqgraph,
  title={Vqgraph: Rethinking graph representation space for bridging gnns and mlps},
  author={Yang, Ling and Tian, Ye and Xu, Minkai and Liu, Zhongyi and Hong, Shenda and Qu, Wei and Zhang, Wentao and Cui, Bin and Zhang, Muhan and Leskovec, Jure},
  journal={arXiv preprint arXiv:2308.02117},
  year={2023}
}

@article{liu2023rethinking,
  title={Rethinking tokenizer and decoder in masked graph modeling for molecules},
  author={Liu, Zhiyuan and Shi, Yaorui and Zhang, An and Zhang, Enzhi and Kawaguchi, Kenji and Wang, Xiang and Chua, Tat-Seng},
  journal={Advances in Neural Information Processing Systems},
  volume={36},
  pages={25854--25875},
  year={2023}
}

@article{wang2024learning,
  title={Learning graph quantized tokenizers},
  author={Wang, Limei and Hassani, Kaveh and Zhang, Si and Fu, Dongqi and Yuan, Baichuan and Cong, Weilin and Hua, Zhigang and Wu, Hao and Yao, Ning and Long, Bo},
  journal={arXiv preprint arXiv:2410.13798},
  year={2024}
}

@article{xia2024opengraph,
  title={Opengraph: Towards open graph foundation models},
  author={Xia, Lianghao and Kao, Ben and Huang, Chao},
  journal={arXiv preprint arXiv:2403.01121},
  year={2024}
}

@article{amiranashvili2024learning,
  title={Learning continuous shape priors from sparse data with neural implicit functions},
  author={Amiranashvili, Tamaz and L{\"u}dke, David and Li, Hongwei Bran and Zachow, Stefan and Menze, Bjoern H},
  journal={Medical Image Analysis},
  volume={94},
  pages={103099},
  year={2024},
  publisher={Elsevier}
}

@article{nan2024hunting,
  title={Hunting imaging biomarkers in pulmonary fibrosis: benchmarks of the AIIB23 challenge},
  author={Nan, Yang and Xing, Xiaodan and Wang, Shiyi and Tang, Zeyu and Felder, Federico N and Zhang, Sheng and Ledda, Roberta Eufrasia and Ding, Xiaoliu and Yu, Ruiqi and Liu, Weiping and others},
  journal={Medical Image Analysis},
  volume={97},
  pages={103253},
  year={2024},
  publisher={Elsevier}
}

@article{stoverud2023aeropath,
  title={AeroPath: An airway segmentation benchmark dataset with challenging pathology},
  author={St{\o}verud, Karen-Helene and Bouget, David and Pedersen, Andre and Leira, H{\aa}kon Olav and Lang{\o}, Thomas and Hofstad, Erlend Fagertun},
  journal={arXiv preprint arXiv:2311.01138},
  year={2023}
}

@article{chu2025deep,
  title={Deep learning-driven pulmonary artery and vein segmentation reveals demography-associated vasculature anatomical differences},
  author={Chu, Yuetan and Luo, Gongning and Zhou, Longxi and Cao, Shaodong and Ma, Guolin and Meng, Xianglin and Zhou, Juexiao and Yang, Changchun and Xie, Dexuan and Mu, Dan and others},
  journal={Nature Communications},
  volume={16},
  number={1},
  pages={2262},
  year={2025},
  publisher={Nature Publishing Group UK London}
}

@article{luo2023efficient,
  title={Efficient automatic segmentation for multi-level pulmonary arteries: The parse challenge},
  author={Luo, Gongning and Wang, Kuanquan and Liu, Jun and Li, Shuo and Liang, Xinjie and Li, Xiangyu and Gan, Shaowei and Wang, Wei and Dong, Suyu and Wang, Wenyi and others},
  journal={arXiv preprint arXiv:2304.03708},
  year={2023}
}

@inproceedings{cheng2024fusion,
  title={Fusion of Machine Learning and Deep Neural Networks for Pulmonary Arteries and Veins Segmentation in Lung Cancer Surgery Planning},
  author={Cheng, Hongyu and Zheng, Limin and Yan, Zeyu and Zhang, Haoran and Meng, Bo and Xu, Xiaowei},
  booktitle={International Conference on Pattern Recognition},
  pages={422--438},
  year={2024},
  organization={Springer}
}

@article{walsh2021imaging,
  title={Imaging intact human organs with local resolution of cellular structures using hierarchical phase-contrast tomography},
  author={Walsh, Claire L and Tafforeau, P and Wagner, WL and Jafree, DJ and Bellier, A and Werlein, C and K{\"u}hnel, MP and Boller, E and Walker-Samuel, S and Robertus, JL and others},
  journal={Nature methods},
  volume={18},
  number={12},
  pages={1532--1541},
  year={2021},
  publisher={Nature Publishing Group US New York}
}

@article{kuo2023terabyte,
  title={Terabyte-scale supervised 3D training and benchmarking dataset of the mouse kidney},
  author={Kuo, Willy and Rossinelli, Diego and Schulz, Georg and Wenger, Roland H and Hieber, Simone and M{\"u}ller, Bert and Kurtcuoglu, Vartan},
  journal={Scientific data},
  volume={10},
  number={1},
  pages={510},
  year={2023},
  publisher={Nature Publishing Group UK London}
}

@article{yagis2023deep,
  title={Deep Learning for Vascular Segmentation and Applications in Phase Contrast Tomography Imaging},
  author={Yagis, Ekin and Aslani, Shahab and Jain, Yashvardhan and Zhou, Yang and Rahmani, Shahrokh and Brunet, Joseph and Bellier, Alexandre and Werlein, Christopher and Ackermann, Maximilian and Jonigk, Danny and others},
  journal={arXiv preprint arXiv:2311.13319},
  year={2023}
}

@article{loshchilov2017decoupled,
  title={Decoupled weight decay regularization},
  author={Loshchilov, Ilya and Hutter, Frank},
  journal={arXiv preprint arXiv:1711.05101},
  year={2017}
}

@inproceedings{peebles2023scalable,
  title={Scalable diffusion models with transformers},
  author={Peebles, William and Xie, Saining},
  booktitle={Proceedings of the IEEE/CVF international conference on computer vision},
  pages={4195--4205},
  year={2023}
}

@article{velivckovic2017graph,
  title={Graph attention networks},
  author={Veli{\v{c}}kovi{\'c}, Petar and Cucurull, Guillem and Casanova, Arantxa and Romero, Adriana and Lio, Pietro and Bengio, Yoshua},
  journal={arXiv preprint arXiv:1710.10903},
  year={2017}
}

@article{rampavsek2022recipe,
  title={Recipe for a general, powerful, scalable graph transformer},
  author={Ramp{\'a}{\v{s}}ek, Ladislav and Galkin, Michael and Dwivedi, Vijay Prakash and Luu, Anh Tuan and Wolf, Guy and Beaini, Dominique},
  journal={Advances in Neural Information Processing Systems},
  volume={35},
  pages={14501--14515},
  year={2022}
}

@InProceedings{Park_2019_CVPR,
author = {Park, Jeong Joon and Florence, Peter and Straub, Julian and Newcombe, Richard and Lovegrove, Steven},
title = {DeepSDF: Learning Continuous Signed Distance Functions for Shape Representation},
booktitle = {Proceedings of the IEEE/CVF Conference on Computer Vision and Pattern Recognition (CVPR)},
month = {June},
year = {2019}
}

@inproceedings{chen2025hierarchical,
  title={Hierarchical Part-Based Generative Model for Realistic 3D Blood Vessel},
  author={Chen, Siqi and Zhang, Guoqing and Lai, Jiahao and Shen, Bingzhi and Zhang, Sihong and Dong, Caixia and Chen, Xuejin and Li, Yang},
  booktitle={International Conference on Medical Image Computing and Computer-Assisted Intervention},
  pages={257--267},
  year={2025},
  organization={Springer}
}

@inproceedings{kuipers2025self,
  title={Self-Supervised Synthetic Cerebral Vessel Tree Generation using Semantic Signed Distance Fields},
  author={Kuipers, Thijs P and Konduri, Praneeta R and Bekkers, Erik J and Marquering, Henk},
  booktitle={Medical Imaging with Deep Learning},
  year={2025}
}

@inproceedings{kuipers2024generating,
  title={Generating cerebral vessel trees of acute ischemic stroke patients using conditional set-diffusion},
  author={Kuipers, Thijs P and Konduri, Praneeta R and Marquering, Henk and Bekkers, Erik J},
  booktitle={Medical Imaging with Deep Learning},
  year={2024}
}

@article{wolterink2018blood,
  title={Blood vessel geometry synthesis using generative adversarial networks},
  author={Wolterink, Jelmer M and Leiner, Tim and Isgum, Ivana},
  journal={arXiv preprint arXiv:1804.04381},
  year={2018}
}

@article{kirst2020mapping,
  title={Mapping the fine-scale organization and plasticity of the brain vasculature},
  author={Kirst, Christoph and Skriabine, Sophie and Vieites-Prado, Alba and Topilko, Thomas and Bertin, Paul and Gerschenfeld, Gaspard and Verny, Florine and Topilko, Piotr and Michalski, Nicolas and Tessier-Lavigne, Marc and others},
  journal={Cell},
  volume={180},
  number={4},
  pages={780--795},
  year={2020},
  publisher={Elsevier}
}

@article{todorov2020machine,
  title={Machine learning analysis of whole mouse brain vasculature},
  author={Todorov, Mihail Ivilinov and Paetzold, Johannes Christian and Schoppe, Oliver and Tetteh, Giles and Shit, Suprosanna and Efremov, Velizar and Todorov-V{\"o}lgyi, Katalin and D{\"u}ring, Marco and Dichgans, Martin and Piraud, Marie and others},
  journal={Nature methods},
  volume={17},
  number={4},
  pages={442--449},
  year={2020},
  publisher={Nature Publishing Group US New York}
}

@article{lipman2022flow,
  title={Flow matching for generative modeling},
  author={Lipman, Yaron and Chen, Ricky TQ and Ben-Hamu, Heli and Nickel, Maximilian and Le, Matt},
  journal={arXiv preprint arXiv:2210.02747},
  year={2022}
}

@inproceedings{zhang2018link,
  title={Link prediction based on graph neural networks},
  author={Zhang, Muhan and Chen, Yixin},
  booktitle={Advances in Neural Information Processing Systems},
  pages={5165--5175},
  year={2018}
}

\clearpage
\appendix

\section{Additional Details on Datasets}

We extract spatial graphs from the ground truth voxel segmentation mask using the Voreen graph extraction tool~\cite{meyer2009voreen} with the bulge size set to three. Our experiments are conducted on the following datasets:
\begin{enumerate}
    \item airway datasets (ATM~\cite{zhang2023multi}, AIIB~\cite{nan2024hunting}, AeroPath~\cite{stoverud2023aeropath})
    \item cerebral vasculature (COSTA~\cite{mou2024costa})
    \item pulmonary vessels (HiPas~\cite{chu2025deep}, PARSE~\cite{luo2023efficient}, Pulmonary-AV~\cite{cheng2024fusion})
\end{enumerate}
\noindent Tab.~\ref{tab:dataset_stats} reports the mean, median, minimum, and maximum numbers of nodes and edges for each dataset. The ranges differ substantially across datasets, indicating variation in graph size. The mean compression ratio ($\kappa$) also varies substantially across datasets, ranging from 4.091 for AeroPath to 14.566 for PARSE.

\begin{figure*}[t!]
    \centering
    \includegraphics[width=\linewidth]{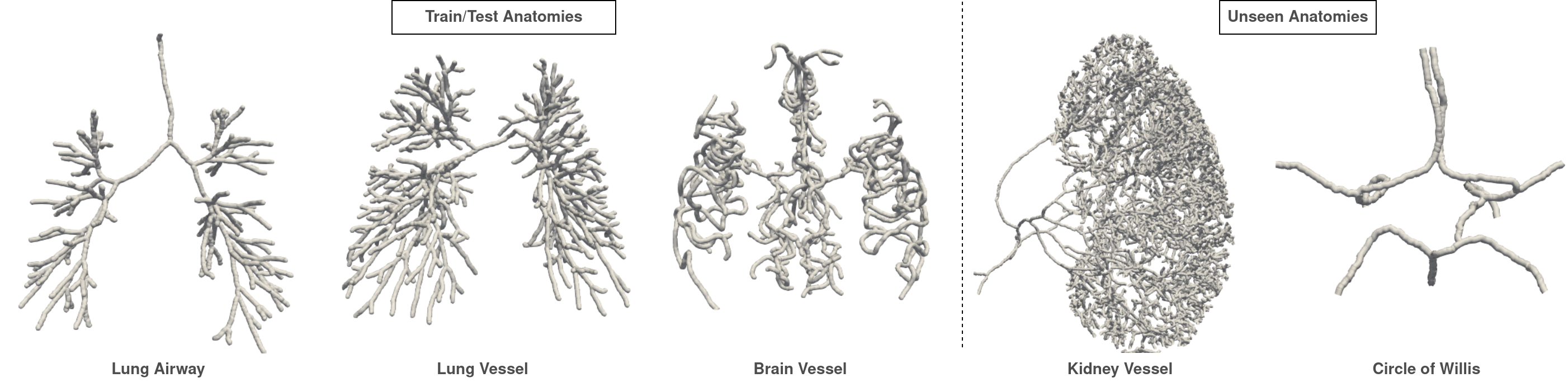}
    \caption{\textbf{Left:} Representative samples from the training distribution. We cover a range of anatomies in our training samples, such as airways, cerebral vasculature, and pulmonary vessels. \textbf{Right:} Representative samples from the renal vasculature and the Circle of Willis dataset used to evaluate the performance of the model on samples not seen during training.}
    \label{fig:dataset_example}
\end{figure*}

\begin{figure*}[t!]
    \centering
    \includegraphics[width=\linewidth]{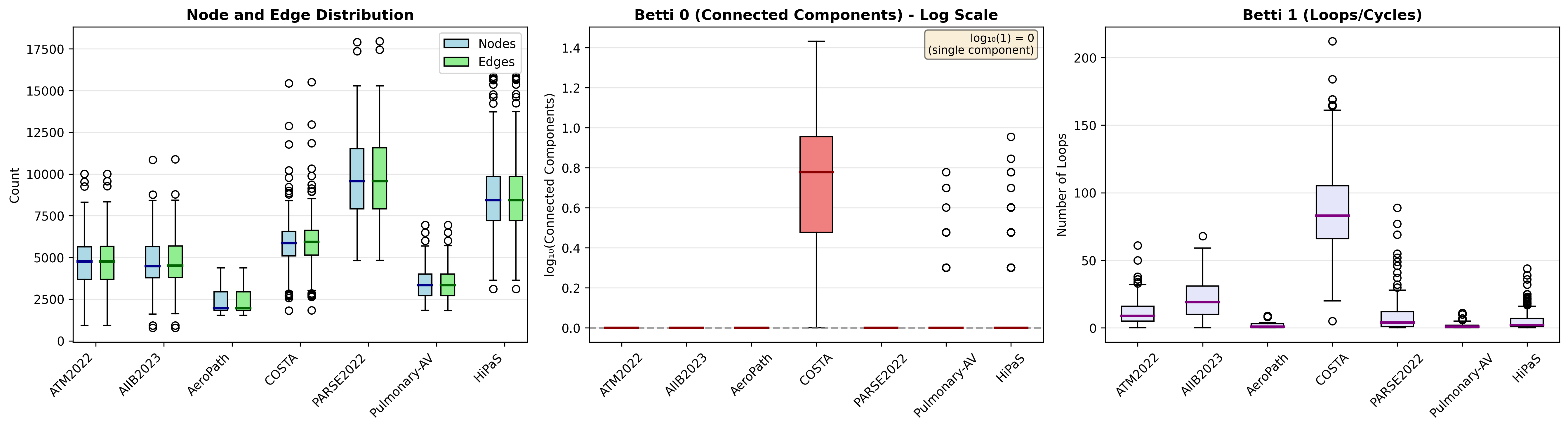}
    \caption{We summarize the structural variability of our datasets by reporting the distributions of node and edge counts across the training set. In addition, we include the number of connected components ($\beta_0$) and loop count ($\beta_1$) for each dataset. The samples exhibit substantial variation in both graph size and topology, with wide ranges in node/edge counts and $\beta_1$. For number of connected components ($\beta_0$), COSTA exhibits maximum variability. This diversity underscores the need for (and the ability of) our model to generalize across anatomies with markedly different scales and topological characteristics.}
    \label{fig:split_stats}
\end{figure*}

Further, across datasets, the $\beta_0$ statistics reveal marked heterogeneity in connectivity. Several airway and pulmonary sets are consistently fully connected, i.e.,  all graphs have a single component (AIIB2023, ATM2022, PARSE2022, AeroPath; 100\% single-component). In contrast, other datasets show fragmentation. HiPaS and Pulmonary-AV are mostly connected (83.6\% and 87.2\% single-component) but can contain up to 9 and 6 components. COSTA is the most fragmented, with only 8.6\% single-component graphs and up to 27 components in extreme cases. The $\beta_1$ differences are even more stark. We see that COSTA samples have more than 200 loops in the extreme case. These patterns highlight that the model must handle both well-connected anatomies and strongly fragmented vascular trees. Fig.~\ref{fig:split_stats} shows the node and edge count variability for the training split along with the number of loops contained in the graphs for each dataset, while Fig.~\ref{fig:dataset_example} includes representative training samples, illustrating substantial heterogeneity in both topology and scale across the dataset.

\begin{table*}[t!]
    \centering
    \scriptsize
    \caption{Summary statistics for each dataset.}
    \label{tab:dataset_stats}
    \begin{tabular}{l c cccc cccc cccc}
        \toprule
        \multirow{2}{*}{\textbf{Split}} & \multirow{2}{*}{$\kappa$} & \multicolumn{4}{c}{\textbf{\# Samples}} & \multicolumn{4}{c}{\textbf{\# Nodes}} & \multicolumn{4}{c}{\textbf{\# Edges}}\\
        \cmidrule(lr){3-6} \cmidrule(lr){7-10} \cmidrule(lr){11-14} 
        &  & Train & Val & Test & Total & Mean & Median & Min & Max & Mean & Median & Min & Max \\
        \midrule
        \textbf{ATM} & 6.759 & 198 & 24 & 43 & 265 & 4614 & 4518 & 915 & 10007 & 4624 & 4520 & 914 & 10016 \\
        \textbf{AIIB} & 7.257 & 57 & 8 & 15 &  80 & 4954 & 4677 & 774 & 10847 & 4974 & 4690 & 773 & 10880 \\
        \textbf{AeroPath} & 4.091 & 13 & 3 & 0 &  16 & 2793 & 2323 & 1535 & 6741 & 2795 & 2330 & 1534 & 6745 \\
        \textbf{COSTA} & 8.597 & 269 & 34 & 52 &  355 & 5869 & 5882 & 1746 & 16088 & 5950 & 5967 & 1760 & 16168 \\
        \textbf{HiPas} & 12.489 & 372 & 47 & 81 &  500 & 8526 & 8348 & 3105 & 15835 & 8529 & 8348 & 3104 & 15846 \\
        \textbf{PARSE} & 14.566 & 73 & 11 & 16 &  100 & 9944 & 9530 & 4813 & 17909 & 9956 & 9533 & 4817 & 17954 \\
        \textbf{Pulmonary-AV} & 5.061 & 164 & 20 & 28 &  212 & 3455 & 3311 & 1576 & 8732 & 3455 & 3313 & 1575 & 8733 \\
        \bottomrule
    \end{tabular}
\end{table*}

\noindent For training, we partition the data into stratified train/validation/test splits (70/10/20), aiming to preserve the distribution of anatomies and graph sizes across splits. Only the training set is used to fit the autoencoders and the downstream diffusion models. The validation set is reserved for hyperparameter selection, while the test set is held out for final reporting. Please note that since the AeroPath dataset has very few samples, it was used only in the training and validation phase.
The final data split has 1146 training samples, 147 validation samples, and 235 test samples. Dataset-specific statistics are presented in Tab.~\ref{tab:dataset_stats}

\begin{figure*}[thb]
    \centering
    \includegraphics[width=\linewidth]{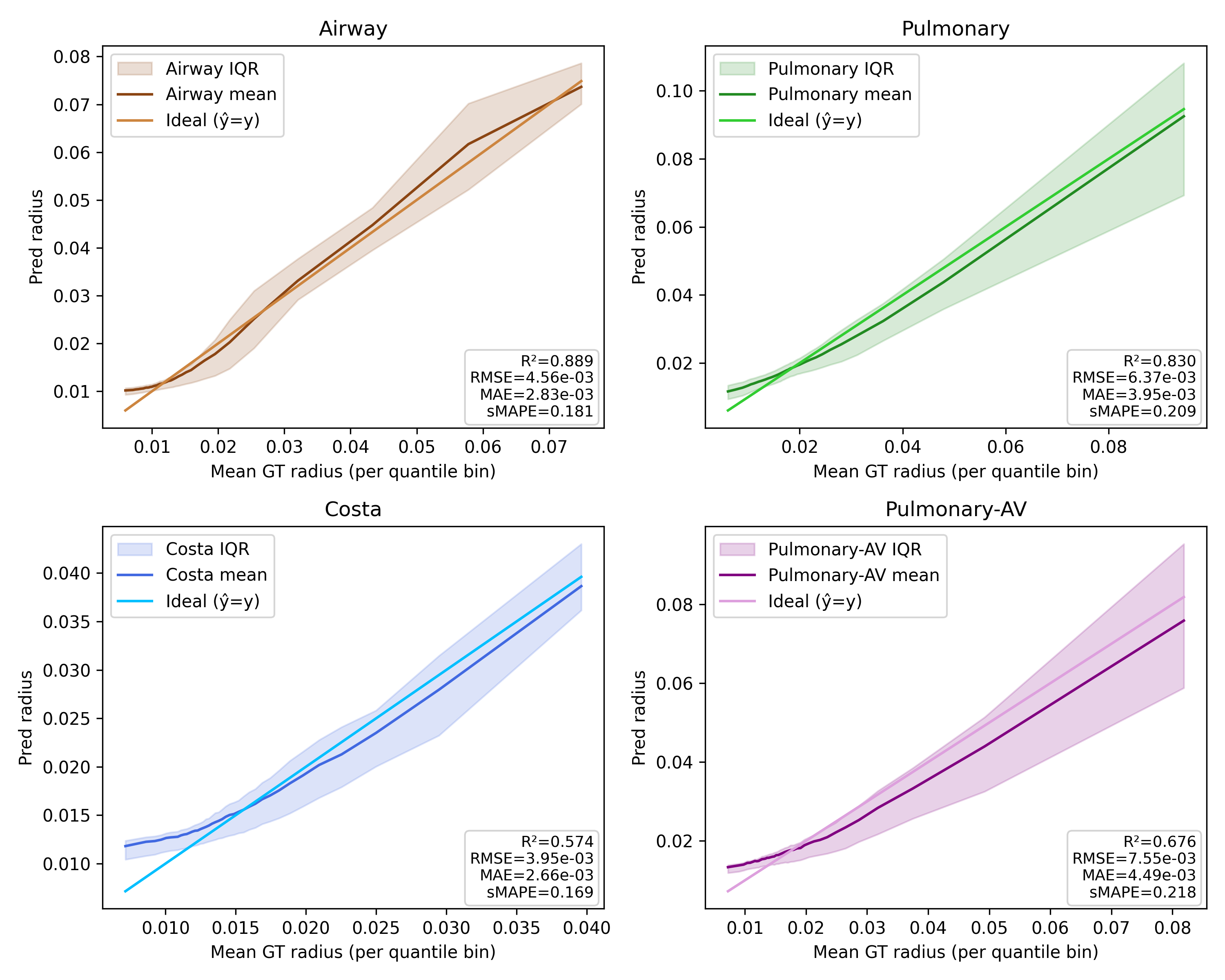}
    \caption{Quantile-binned reliability diagram for edge-radius regression across anatomical categories. Edges are grouped into quantile bins by ground-truth radius (x-axis). For each bin we plot the mean predicted radius (y-axis) with the interquartile range (shaded). The identity line ($\hat r=r$) indicates perfect agreement between predictions and ground truth.}
    \label{fig:calibration_performance}
\end{figure*}

\section{Predicting Radius From Structure}\label{str_from_pred}
In this work, we prioritize accurate prediction of graph topology and therefore assign a pseudo-radius to each graph edge. 
This design simplifies learning by avoiding heterogeneous radius variations during structure prediction. We hypothesize that reliable radii can be recovered once the graph structure is known via a separate and straight-forward post hoc step. Accordingly, we train a lightweight edge-radius regression model for each of the four anatomical categories. For this, we use the same train-val-test split introduced in supplementary sec.~A. The model is parameterized by a transformer encoder with 4 layers, 6 attention heads, and a hidden dimension of 384. The model takes node coordinates and edge connectivity as input. Normalized node coordinates are embedded with a Fourier positional encoding and processed by the encoder to produce node features. Edge radii are predicted by concatenating the features of the two incident nodes and passing them through an MLP.

To handle the skewed, heavy-tailed distribution of \emph{edge} radii, we optimize a quantile-balanced objective: edges are stratified by ground-truth radius quantiles $(q50, q90, q99)$, and we macro-average a Huber loss on the log-relative error $\log(1+\hat r)-\log(1+r)$ across bins to prevent the abundant small-radius edges from dominating training. We assess agreement between predicted and ground-truth radii using \emph{quantile-binned reliability diagrams} (Fig.~\ref{fig:calibration_performance}), which plot the mean predicted radius (with interquartile range) against the mean ground-truth radius per quantile bin, together with the identity line $\hat r=r$. The regression model reliably predicts radii, achieving $R^2$ values of 0.889 (airways), 0.574 (COSTA), 0.830 (full pulmonary), and 0.676 (Pulmonary-AV), with corresponding sMAPEs of 0.181, 0.169, 0.209, and 0.218. While the $R^2$ value for COSTA is lower than for the other datasets, it should be interpreted in context: $R^2$ is computed relative to a constant mean predictor and is strongly influenced by the variance of the target distribution, such that datasets with a narrower spread of radii can yield lower $R^2$ even when absolute errors are comparable. In addition, COSTA radius annotations are noisier, and we do not apply denoising, such as a median filter, to refine them. For cross-dataset comparison, we therefore also report sMAPE, which normalizes errors by the magnitudes of the prediction and target and is less sensitive to differences in radius scale and spread. These results indicate that the regression model maintains consistently low relative error across anatomical categories, thereby implicitly validating our design choice to shift the focus to learning vascular structure and predicting the radius as a post-hoc step. We stress that our design choice does not trivialize the problem, but instead shifts the focus to structural fidelity (encoding global, 3D geometry and topology), while decoupling it from the severe scale imbalance caused by wide radius ranges.

\section{Metrics}
\subsection{FID Computation}
To compute FID, we train a PointNet++ classifier~\cite{qi2017pointnet++} to predict anatomical class labels using only raw centerline points (no surface normals). Inputs are normalized to [-1, 1] and centered to zero mean. The network comprises four Set Abstraction (SA) layers and is trained for 100 epochs on the training split. We select the best model based on validation loss. After training, we discard the classification head and use the penultimate backbone features as embeddings, yielding a 1024-dimensional representation per sample. FID is then evaluated between the Gaussian fits of these embeddings for generated versus training centerlines.

\subsection{Centerline Dice (clDice)}
The centerline Dice (clDice) metric~\cite{shit2021cldice} is a topology-preserving metric of tubular structures by aligning predicted centerlines with ground-truth masks (and vice versa). Let \(\Omega_\tau \in \{0,1\}^{|\mathcal{X}|}
\) be a predicted discretized occupancy field and \(\Omega \in \{0,1\}^{|\mathcal{X}|}\) a ground-truth discretized occupancy field over domain \(\mathcal{X}\). We obtain centerline \(\mathcal{S}(\cdot)\) on this discretized domain as follows:
\begin{equation}
  V = \mathcal{S}(\Omega), \qquad \hat{V} = \mathcal{S}(\Omega_{\tau}).
\end{equation}
Topology precision and sensitivity are defined as
\begin{equation}
\begin{aligned}
  T_{\mathrm{sens}} &= \frac{\langle V,\, \Omega_\tau\rangle + \varepsilon}{\langle V,\, \mathbf{1}\rangle + \varepsilon} \quad \text{and} \\
  T_{\mathrm{prec}} &= \frac{\langle \hat{V},\, \Omega\rangle + \varepsilon}{\langle \hat{V},\, \mathbf{1}\rangle + \varepsilon},
\end{aligned}
\end{equation}
where \(\langle A,B\rangle = \sum_{x \in \mathcal{X}} A(x)\,B(x)\) denotes the point-wise inner product, \(\mathbf{1}\) is the all-ones map, and \(\varepsilon > 0\) ensures numerical stability. The clDice score is the harmonic mean
\begin{equation}
  \text{clDice}(\Omega,\Omega_\tau) = \frac{2\, T_{\mathrm{prec}} \, T_{\mathrm{sens}}}{T_{\mathrm{prec}} + T_{\mathrm{sens}}}.
\end{equation}

\noindent Please note that in our experiments, clDice is used only as an evaluation metric. It quantifies whether the reconstructed occupancy field preserves the centerline.

\subsection{Betti Error}
The Betti error measures topological consistency between a reconstructed graph and its reference by comparing their Betti numbers. For graphs, \(\beta_{0}\) counts \emph{connected components} while \(\beta_{1}\) counts \emph{independent loops}. Given a reconstruction \(\hat{G}\) and ground truth \(G\), we define:
\begin{equation}
\begin{aligned}
|\Delta \beta_0| \;&=\; \bigl|\,\beta_{0}(\hat{G}) - \beta_{0}(G)\,\bigr| \;\; \text{and} \\
|\Delta \beta_1| \;&=\; \bigl|\,\beta_{1}(\hat{G}) - \beta_{1}(G)\,\bigr|.
\end{aligned}
\end{equation}

\noindent A lower value of $|\Delta \beta_0|$ and $|\Delta \beta_1|$ indicates better preservation of global connectivity (\(\beta_{0}\)) and loop structure (\(\beta_{1}\)) respectively. We compute the Betti numbers on the graph extracted from the occupancy network following the procedure outlined in Section~3.2.
Unlike geometric metrics (\eg, Chamfer Distance) that reward pointwise proximity, Betti errors are relatively insensitive to small spatial perturbations yet penalize topological defects such as broken branches (increased \(\beta_{0}\)) or spurious/missing loops (changes in \(\beta_{1}\)). We therefore report Betti error alongside clDice and CD to assess topology preservation and geometric fidelity.

\section{Implementation Details}
\subsection{Model Architecture}
We follow the VAE design of Zhao et al.~\cite{zhao2025hunyuan3d}. Given an input graph $G=(V,E,\mathbf{P})$ with centerline points $\mathbf{P}=\{p_i\}_{i=1}^{N}\subset\mathbb{R}^3$, we first apply farthest‐point sampling (FPS)~\cite{qi2017pointnet++} to select query points $\mathbf{Q_{in}}=\{q_j\}_{j=1}^{K}$. Features are aggregated from the full set $\mathbf{P}$ to the query points $\mathbf{Q_{in}}$ via a cross-attention layer. The resulting query points are processed by an 8-layer self-attention encoder that outputs the VAE parameters (mean $\mu$ and log-variance $\log\sigma^2$) for the latent $z\in\mathbb{R}^{K\times C}$. Sampling uses the reparameterization trick $z=\mu+\sigma\odot\epsilon$. A 16-layer decoder then maps $z$ to a continuous graph occupancy field.

\begin{figure*}[t!]
    \centering
    \includegraphics[width=0.8\linewidth]{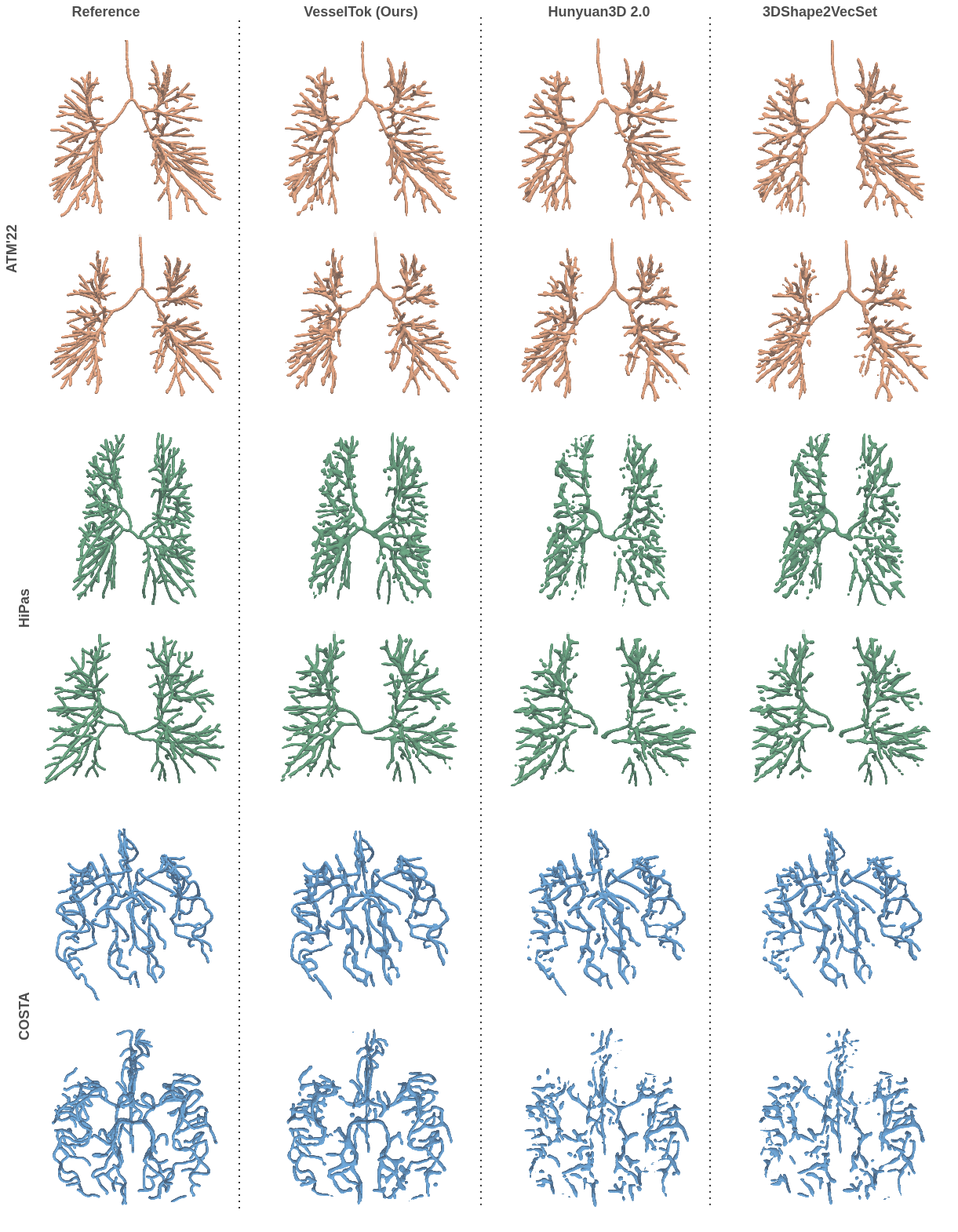}
    \caption{Additional qualitative results for the graph reconstruction task. We find that VesselTok demonstrates superior reconstruction capabilities.}
    \label{fig:supp_qual}
\end{figure*} 

\subsection{Training Details}
We train the VAE for 24,000 epochs using 2,048 query points per sample to compute the occupancy loss. To bias supervision toward the vessel surface, 50\% of query points are sampled near the centerlines by adding Gaussian perturbations with $\sigma \in [0.005, 0.05]$. The remaining query points are drawn from the ambient volume. We train the model using a peak learning rate of $5\times10^{-5}$, followed by cosine decay to a minimum learning rate of  $1\times10^{-6}$. The model is trained on a single GPU (NVIDIA A100 GPU, 80 GB VRAM) with batch size 32, requiring $\sim4$ days. To handle variable-size input graphs, sequences are padded to the per-mini-batch maximum node count, and padded positions are excluded from loss computation.
Inference per sample takes $\sim$9.65\,s for our VesselTok, compared to $\sim$9.13\,s (Hunyuan3D 2.0) and $\sim$9.05\,s (3DShape2VecSet). Peak VRAM usage is $\sim$24\,GB across all methods.

\subsection{Baselines}
We compare our method against Hunyuan3D 2.0~\cite{zhao2025hunyuan3d} and 3DShape2VecSet~\cite{zhang20233dshape2vecset}. Both baselines operate on surface points, whereas our method operates on centerlines.
3DShape2VecSet employs a closely related VAE design, using a single cross-attention layer to produce the VAE parameters and a 24-layer decoder over FPS-selected query points $\mathbf{Q_{in}}$ to predict the occupancy field. Our architecture is most similar to Hunyuan3D 2.0. We adopt an 8-layer encoder to estimate the VAE parameters, along with a decoder of comparable depth. Hunyuan3D 2.0 also proposes curvature-aware sampling for query points selection. In our experiments, this strategy offered no measurable advantage, so we used standard farthest-point sampling (FPS) to select the query points.

\subsection{Hyperparameters}
We employ $\lambda=1e^{-3}$ for Eq. (2) and $\tau=0.5$ for predicting the occupancy value. For rendering the 512$^3$ grid size during inference, we use a chunk size of 64$^3$. We use random rotations as a data augmentation on the graphs to improve SE(3) robustness. We train the model using the AdamW~\cite{loshchilov2017decoupled} optimizer.

\begin{figure*}[t!]
    \centering
    \includegraphics[width=0.8\linewidth]{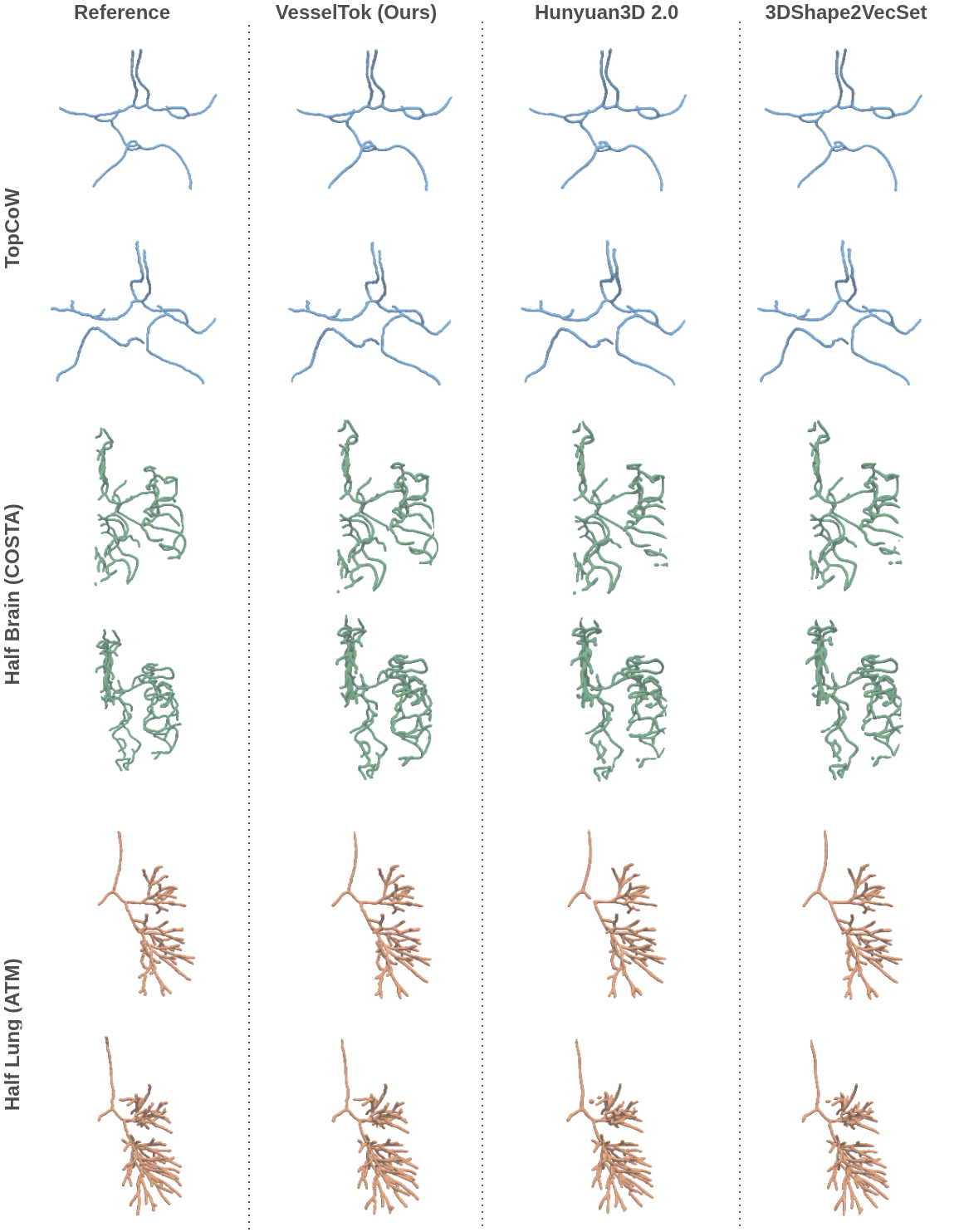}
    \caption{Additional qualitative results for the graph reconstruction task of previously unseen domains. Our results demonstrate VesselTok's strong prior, resulting in robust reconstructions.}
    \label{fig:ood_supp}
\end{figure*} 

\subsection{Ablation Setup}
We perform ablation experiments on two key design choices: (i) the pseudo-radius \(r\) used to construct the occupancy field, and (ii) the VAE configuration, namely the number of query points \(K\) and latent channel dimension \(C\). For both studies, all models are trained on the ATM dataset~\cite{zhang2023multi} for \(16{,}000\) epochs with a batch size of \(32\) and a cosine-annealed learning rate schedule from \(5\times 10^{-5}\) down to \(1\times 10^{-6}\) with AdamW~\cite{loshchilov2017decoupled} optimizer. Experiments are conducted on a single GPU (NVIDIA A100 GPU, 80 GB VRAM) and require $\sim$24 hours per run. We apply random rotations, flips, and slight translations to the samples during training as data augmentation. The ATM dataset is split into train/validation/test sets in a 70/10/20 ratio, and all ablation results are reported on the held-out test split.

\section{Additional Reconstruction Visualization}

Fig.~\ref{fig:supp_qual} illustrates additional qualitative comparisons between Hunyuan3D 2.0, 3DShape2VecSet, and VesselTok. Across diverse anatomies and graph sizes, VesselTok more faithfully preserves thin peripheral branches and long-range connectivity, while producing fewer spurious fragments or missed segments than the baselines. In contrast, 3DShape2VecSet and Hunyuan3D 2.0 more often exhibit discontinuities and occasional topological artifacts.

\section{Additional Results on Unseen Anatomies}
We evaluate VesselTok’s generalization beyond the training distribution. Specifically, we test renal vasculature as an extreme scalability case and sagittal-plane clipped half-lung and half-brain samples to assess robustness to structural perturbations. We describe these experiments in more detail below.

\subsection{Renal Vasculature}
The renal vasculature provides an extreme test of generalizability. The graphs are approximately one order of magnitude larger than those seen during training. To make inference tractable in this setting, we adopt a chunked reconstruction strategy. Specifically, we partition the full-resolution volume into fixed-size crops of $150^3$ voxels. For each crop, we extract the subset of the centerline graph lying inside the crop, translate it to its center of mass (without any scaling), and run the VAE-based reconstruction on this localized subgraph. The reconstructed crops are then mapped back to their original coordinate frames and aggregated to obtain a full-volume reconstruction. No additional alignment or blending is applied during the patching step. As a light post-processing step, we remove small isolated components by discarding any connected component with fewer than 100 voxels. As illustrated in Fig.~\ref{fig:ood_supp}, our model reconstructs the renal vasculature more faithfully than the baselines.

\subsection{Unseen Anatomies (Half-Lung and Half-Brain)} To assess the generalizability of our method to anatomies not seen during training, we construct a synthetic dataset by applying sagittal-plane clipping to samples from both ATM and COSTA. Specifically, each anatomy is intersected with the sagittal plane and only one side is retained, producing deliberately extreme, anatomically implausible cases. We refer to the resulting samples as \emph{half-lung} and \emph{half-brain}. We use this terminology to denote strongly truncated anatomies obtained via sagittal clipping. These samples probe the model's ability to maintain reconstruction quality under severe structural perturbations. To avoid data leakage, we generate them exclusively from the test splits of the original datasets and evaluate only the reconstruction performance of the VAE encoder $\mathcal{T}$ on this held-out, synthetically perturbed set.

\section{Additional Details on Generative Model}
In this section, we detail the training procedure for our generative model, including architectural choices, the training objective, and implementation details. We also present qualitative results. Visually, VesselTok generates anatomies that more closely follow the underlying data distribution, whereas baseline methods exhibit more topological errors. We discuss the experimental setup and results in more detail below.

\subsection{Latent Diffusion for Anatomical Graph Synthesis}
Following~\cite{zhang20233dshape2vecset}, we train an Elucidated Diffusion Model (EDM)~\cite{karras2022elucidating} to generate both category-conditioned and unconditional samples, with conditioning defined by anatomical labels. For this, we define four categories:
\begin{enumerate}
  \item Airway: all samples from ATM, AIIB, and AeroPath datasets
  \item Costa: samples (cerebral vasculature), separated due to distinct brain anatomy
  \item Full-Pulmonary: samples from HiPas and PARSE 
  \item Pulmonary-AV: contains only one vascular subsystem (arterial or venous) per case
\end{enumerate}
The anatomical category is encoded as a learnable embedding and used to condition the generative model. During training, we randomly drop the category embedding in 25\% of the batches, which encourages the model to also support unconditional generation.\\

\begin{figure*}[t!]
    \centering
    \includegraphics[width=0.95\linewidth]{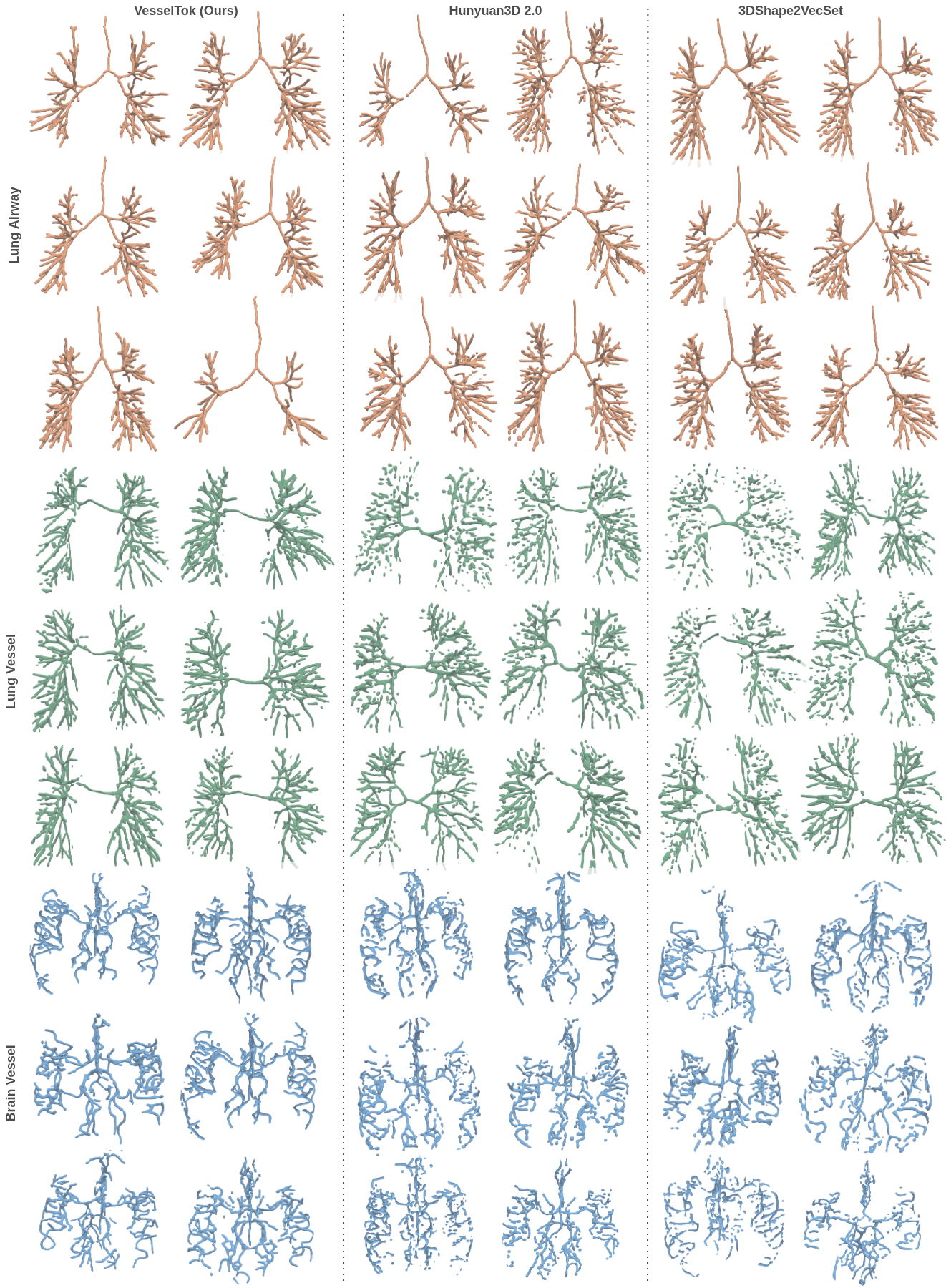}
    \caption{Additional qualitative results for conditional generation. VesselTok consistently generates more realistic vessels in comparison to previous methods.}
    \label{fig:condgene_supp}
\end{figure*}

\noindent\textbf{Training setup.}
We adopt a latent diffusion pipeline. At each step, centerline points are encoded by the VAE into a fixed-length latent, and the EDM is trained directly in this latent space. The EDM backbone uses 24 self-attention blocks. We train for 16{,}000 epochs with a 800 epoch warm-up that linearly increases the learning rate from \(1\!\times\!10^{-6}\) to \(1\!\times\!10^{-4}\), followed by a fixed rate of \(1\!\times\!10^{-4}\) thereafter. The batch size is set to 32. Training requires $\sim$3 days. The approach is diffusion-backbone agnostic, and in principle, the EDM can be replaced with alternative generative solvers (\eg, flow matching) without modifying the tokenization stage.\\

\noindent\textbf{Sampling and evaluation.}
For conditional generation, we sample 500 graphs per category, while for unconditional generation, we draw {1{,}000} samples. The sampling hyperparameters follow the default settings from~\cite {karras2022elucidating}.

\subsection{Qualitative Results}
Fig.~\ref{fig:condgene_supp} presents qualitative examples of conditional samples generated by our model compared with the baseline. Visually, VesselTok produces anatomies that more closely follow the underlying data distribution, whereas the baseline methods exhibit higher topological errors. These examples complement the quantitative results of our generated graphs.

\begin{figure}[thb]
    \centering
    \includegraphics[width=0.8\linewidth]{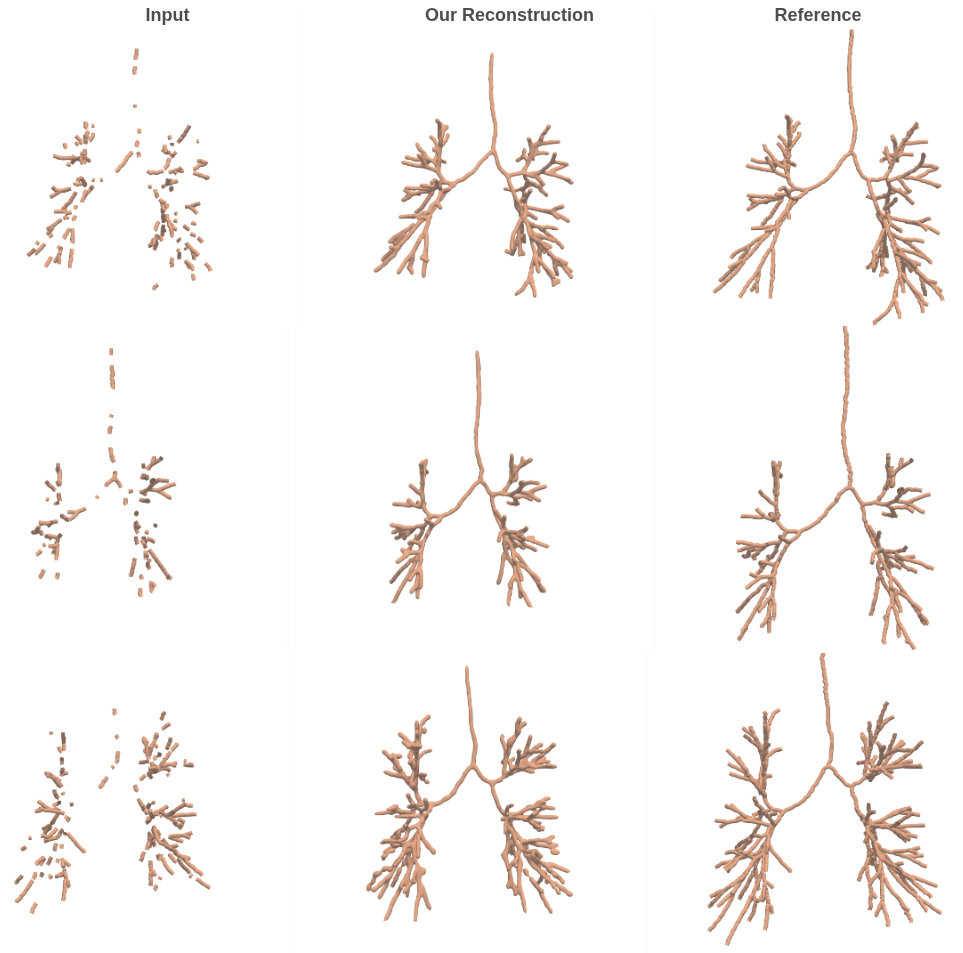}
    \caption{Qualitative examples from our generative model based link prediction on airway dataset.}
    \label{fig:link_pred}
\end{figure}

\section{Link Prediction via Conditional Flow Matching}
We describe the link-prediction (infill) task on the airway tree (ATM) dataset. We first outline how incomplete graphs are constructed and detail the training setup and architecture of the conditional model. We then summarize the baseline implementation used for comparison. Finally, we present qualitative examples illustrating VesselTok’s inpainting performance.
\subsection{Setup}
We tackle the link prediction (infill) task using a conditional flow-matching model. To construct training pairs, we generate a synthetic partial-observation dataset from each point cloud \(\mathbf{P}\). Concretely, we randomly select \(5\%\) of points \(p \in \mathbf{P}\) and, for each selected point, remove all nodes in its \(20\)-hop neighborhood, yielding a partial input \(\tilde{\mathbf{P}}\). This procedure removes $\sim$40\% of spatially adjacent edges. For validation and test samples, the edge mask is fixed, whereas for training samples, the edge removal is resampled on each iteration. This partial point cloud serves as the conditioning signal. For training, both the original point cloud \(\mathbf{P}\) and its partial counterpart \(\tilde{\mathbf{P}}\) are encoded with the same encoder $\mathcal{T}$. We use a DiT-style architecture to implement conditional flow matching on the latent representation. At inference time, only the incomplete point cloud is available, and the model infills the missing regions based on the learned prior. We train the infill model exclusively on the training split. For evaluation, we fix a random seed, generate partial observations \(\tilde{\mathbf{P}}\) from the test split using the same masking protocol, and report performance on these held-out incomplete graphs.\\

\noindent\textbf{Training details.}
We perform the link prediction task on the ATM split only, using 198 training samples. The conditional flow-matching model is trained for \(\,10{,}000\) epochs with an \(\,800\)-epoch warm-up. The learning rate is linearly increased from \(1\times 10^{-6}\) to \(1\times 10^{-4}\) during warm-up and then kept fixed at \(1\times 10^{-4}\) for the remaining epochs. As backbone, we adopt the DiT architecture from~\cite{peebles2023scalable}, configured with \(16\) double-stream blocks and \(8\) single-stream blocks~\cite{zhao2025hunyuan3d}. During training, we apply a random 3D rotation to each point-cloud sample for data augmentation. The model is trained with a batch size of \(16\) on a single GPU (NVIDIA A100 GPU, 80 GB VRAM), and training completes in approximately two days.

\subsection{Implicit Neural Representation (INR) Baseline}
As an additional, non-generative baseline, we employ a DeepSDF-style auto-decoder~\cite{amiranashvili2024learning,Park_2019_CVPR} that represents each vessel tree as a continuous signed distance field (SDF). The model associates every train volume with a learnable latent vector of dimension $1024$. Given a query coordinate $\mathbf{x} \in \mathbf{R}^{3}$ and the corresponding latent code, an implicit neural representation predicts the SDF value using an 8 layer multilayer perceptron with a hidden dimension of $256$. We adopt sinusoidal representation networks (SIREN)~\cite{sitzmann2020implicit} as activation functions to better capture high-frequency detail, using the recommended first-layer frequency $\omega_{0} = 30 $. No additional positional encoding is applied.

\subsection{Qualitative Results}
We provide qualitative examples in Fig.~\ref{fig:link_pred}. Given an incomplete input, our method successfully reconstructs missing links, yielding vascular trees that are visually consistent and topologically coherent. These results illustrate that the model has learned a strong vascular prior that can be effectively leveraged for the link-prediction task.

\section{Additional Ablation using Graph Structure}
In all experiments, we work with vessel centerlines represented as graphs with nodes and edges. The connectivity is explicitly used only when constructing the occupancy field for training (Section~3.1). The encoder and decoder themselves operate purely in the point-cloud domain. This naturally raises the question of whether we should exploit the underlying graph structure more directly, \eg, by replacing point-based architectures with graph neural networks (GNNs). In particular, one may ask whether using GAT-Conv~\cite{velivckovic2017graph} or a GraphTransformer~\cite{rampavsek2022recipe} on the centerline graph is beneficial compared with the point-based Transformer design of~\cite{zhao2025hunyuan3d}. We investigate this through three variants:\\

\noindent\textbf{Graph-derived positional encodings (PE).}
We compute graph-based positional encodings for the original nodes \(\mathbf{P}\), namely the top \(20\) Laplacian eigenvectors and \(20\) random-walk eigenvectors. These graph-derived features are concatenated with standard Fourier positional encodings for each point in \(\mathbf{P}\), providing an indirect injection of graph structure via node features, while keeping the architecture otherwise unchanged.\\

\noindent\textbf{GNN on query points (GNN).}
Instead of treating the FPS-selected query points $\mathbf{Q_{in}}$ as an unordered set, we endow them with an explicit graph structure by adding edges based on a depth-first traversal of the original centerline graph. We then replace the Transformer self-attention blocks that act on the query points with GAT-Conv layers, thereby performing message passing over the induced query point graph.\\

\noindent\textbf{GraphTransformer on query points (GT).}
Using the same induced query point graph as above, we replace GAT-Conv with a GraphTransformer \cite{rampavsek2022recipe}, allowing attention-based message passing constrained by the learned adjacency structure.\\

\noindent To quantify the impact of incorporating explicit graph structure, we train all variants for \(10{,}000\) epochs on the ATM dataset~\cite{zhang2023multi} with batch size \(32\), cosine-annealed learning rate from \(5\times 10^{-5}\) down to \(1\times 10^{-6}\), and identical optimization hyperparameters. Table~\ref{tab:graph_str} summarizes the results. Across metrics, we do not observe a consistent advantage for any of the graph-augmented variants over the point-based baseline. One plausible explanation is that standard self-attention already performs global message passing on a fully connected graph, capturing long-range interactions that are critical for these anatomical structures. Thus, explicitly incorporating the graph structure does not seem to be critical. To keep the method simple, we therefore adopt the point-cloud formulation and forego explicit graph-processing layers in our final model.

\begin{table}[!t]
\centering
\caption{Effect of incorporating graph structure on the reconstruction quality. We do not see a clear advantage of explicitly incorporating the graph structure, and hence, we adopt a point cloud formulation instead.}
\begin{tabular}{c|c c c c }
\toprule
Method & clDice $\uparrow$ & CD $\downarrow$ & $|\Delta \beta_0|\downarrow$ & $|\Delta \beta_1|\downarrow$ \\
\midrule
PE & 96.88 & \textbf{0.055} & 0.878 & 9.19 \\
GNN  & 96.27 & 0.057 & 0.890 & 8.90 \\
GT  & 96.36 & 0.056 & \textbf{0.875} & 9.30 \\
Ours  & \textbf{96.95} & 0.061 & 0.881 & \textbf{8.75} \\
\bottomrule
\end{tabular}
\label{tab:graph_str}
\end{table}

\section{Additional Performance Analysis}

\noindent\textbf{Uncertainty and significance.} We verify statistical robustness using multiple random seeds and a paired Wilcoxon test. On the ATM dataset, three runs with different seeds show low run-to-run variance in clDice (0.028). A paired Wilcoxon test against each baseline indicates that VesselTok yields statistically significant improvements on average ($p<0.001$).\\

\noindent\textbf{Sensitivity to Threshold, Skeletonization, \& Grid Resolution}
We evaluate the robustness of our graph extraction pipeline on the ATM dataset by varying three key preprocessing choices: binarization threshold, skeletonization method, and grid resolution. As shown in Tab.~\ref{tab:sensitivity}, performance is stable across thresholds from 0.40 to 0.60, with clDice remaining nearly unchanged around the default threshold of 0.5 and replacing our skeletonization procedure with Voreen results in minimal change in clDice. Finally, Betti errors remain consistent across grid resolutions of 512, 640, and 768, indicating that the extracted graph topology is not strongly sensitive to the chosen discretization. Overall, these results suggest that our evaluation pipeline is robust to reasonable variations in thresholding, skeletonization, and resolution. \\

\begin{table}[!h]
    \centering
     \caption{We evaluate the effect of binarization threshold, skeletonization method, and grid resolution on ATM.}
     \label{tab:sensitivity}
    \begin{tabular}{ccccc|c|ccc|ccc}
    \hline
        \multicolumn{5}{c|}{clDice $\uparrow$} & clDice $\uparrow$ 
        & \multicolumn{3}{c|}{$|\Delta \beta_0|\downarrow$} 
        & \multicolumn{3}{c}{$|\Delta \beta_1|\downarrow$}
         \\
        \cline{1-5}\cline{7-12}
        \multicolumn{5}{c|}{Thresholds  (Resolution=512)} & Voreen 
        & \multicolumn{6}{c}{Resolutions (Threshold=0.5)} 
        \\
        \cline{1-5}\cline{7-12}
        0.40 & 0.45 & 0.50 & 0.55 & 0.60 & skeleton 
        & 512 & 640 & 768 
        & 512 & 640 & 768 
        \\
        \hline
        96.84 & 96.87 & 96.94 & 96.94 & 96.93 & 96.87
        & 0.08 & 0.09 & 0.11
        & 9.34 & 9.04 & 8.87
         \\
        \hline
    \end{tabular}
\end{table}

\noindent\textbf{Link Prediction Baseline:} We compare against two link-prediction baselines for the graph infill task. A heuristic Jaccard-based method and the GNN-based SEAL model~\cite{zhang2018link}. This task is highly ill-posed, as multiple plausible edge completions may exist for the same partially observed graph, but only some preserve the correct vascular topology. As shown in Tab.~\ref{tab:link_pred}, the Jaccard baseline fails to recover meaningful topology (resulting in large Betti errors). SEAL performs substantially better, but still introduces notable errors in both connected components and loops. In contrast, our method achieves the lowest $|\Delta\beta_0|$ and $|\Delta\beta_1|$, indicating more faithful recovery of vascular connectivity and loop structure. \\
\begin{table}[!h]
    \centering
    \caption{Link prediction baseline comparison.} 
    \label{tab:link_pred}
    \begin{tabular}{c|cc}
        \hline
        Method & $|\Delta \beta_0|\downarrow$ & $|\Delta \beta_1|\downarrow$ \\
        \hline
        Jaccard & 2936.26 & 1437.19 \\
        SEAL    & 10.62   & 53.26  \\
        VesselTok (Ours)     & \textbf{3.19} & \textbf{8.58} \\
        \hline
    \end{tabular}
\end{table}

\noindent\textbf{Domain Realism:}
We compute the 1-Wasserstein distance ($W_1$) between real and conditionally generated graphs for scalar properties including edge length ($E_L$), edge angle ($E_{\angle}$), node degree ($Deg$), and tortuosity ($T$). As shown in Tab.~\ref{tab:domain_realism}, VesselTok achieves the lowest distances for edge length, node degree, and tortuosity, while remaining competitive on edge angle.
\begin{table}[!h]
    \centering
    \caption{We evaluate conditionally generated graphs using the 1-Wasserstein distance ($W_1$) between real and generated distributions of scalar domain-specific properties: edge length ($E_L$), edge angle ($E_{\angle}$), node degree ($Deg$), and tortuosity ($T$). Lower values indicate closer agreement with real vascular morphology.}
    \label{tab:domain_realism}
    \begin{tabular}{c|cccc}
    \hline
        Method & $|E_{L}|\downarrow$\; & \;$|E_{\angle}|\downarrow$\; & \;$|Deg|\downarrow$\; & \;$|T|\downarrow$\;\\
        \hline
        Hunyuan3D 2.0   & 0.0231 & \textbf{1.215}  & 0.027 & 0.251 \\
        3DShape2VecSet & 0.0212 & 1.322 &0.026 & 0.236 \\
        VesselTok (Ours) & \textbf{0.0205} & 1.330  & \textbf{0.017} & \textbf{0.054}\\
        \hline
    \end{tabular}
\end{table}

\end{document}